\documentclass{article}
 \PassOptionsToPackage{numbers, compress,sort}{natbib}
 \usepackage[final]{neurips_2020}
\usepackage[ruled,vlined]{algorithm2e}
\usepackage{algorithmic}
\usepackage[utf8]{inputenc} % allow utf-8 input
\usepackage[T1]{fontenc} % use 8-bit T1 fonts
\usepackage[hyperfootnotes=false]{hyperref} % hyperlinks
\usepackage{url} % simple URL typesetting
\usepackage{booktabs} % professional-quality tables
\usepackage{amsfonts} % blackboard math symbols
\usepackage{nicefrac} % compact symbols for 1/2, etc.
\usepackage{microtype} % microtypography
\usepackage[utf8]{inputenc} % allow utf-8 input
\usepackage[T1]{fontenc} % use 8-bit T1 fonts
\usepackage{hyperref} % hyperlinks
\usepackage{url} % simple URL typesetting
\usepackage{booktabs} % professional-quality tables
\usepackage{amsfonts} % blackboard math symbols
\usepackage{nicefrac} % compact symbols for 1/2, etc.
\usepackage{microtype} % microtypography
\usepackage{graphicx}
\usepackage[table]{xcolor}
\usepackage{gensymb}
\usepackage{subcaption}
\usepackage{amsthm}
\usepackage{mathrsfs}
\usepackage{amsmath}
\usepackage{amssymb}
\usepackage{mathtools}
\usepackage{wrapfig}
\usepackage{soul}
\makeatletter
\def\BState{\State\hskip-\ALG@thistlm}
\makeatother

\newtheorem{theorem}{Theorem}

\newtheorem{lemma}[theorem]{Lemma}

\usepackage[textwidth=1.2in,textsize=tiny]{todonotes}
\newcommand{\E}{\mathbb{E}}
\newcommand{\bas}[1]{\begin{align*}#1\end{align*}}
\newcommand{\phiv}{\boldsymbol{\phi}}
\newcommand{\ba}[1]{\begin{align}#1\end{align}}

\newcommand{\minus}{\scalebox{0.5}[1.0]{$-$}}

\usepackage{stackengine}

\makeatletter
\newcommand{\distas}[1]{\mathbin{\overset{#1}{\kern\z@\sim}}}%

\newcommand{\beqs}{\vspace{0mm}\begin{eqnarray}}
\newcommand{\eeqs}{\vspace{0mm}\end{eqnarray}}
\newcommand{\barr}{\begin{array}}
\newcommand{\earr}{\end{array}}

\newcommand{\av}[0]{{\boldsymbol{a}}}

\newcommand{\hv}[0]{{\boldsymbol{h}} }

\newcommand{\pv}[0]{{\boldsymbol{p}}}

\newcommand{\xv}{\boldsymbol{x}}
\newcommand{\yv}{\boldsymbol{y}}

\newcommand{\alphav}{\boldsymbol{\alpha}}

\newcommand{\epsilonv}{\boldsymbol{\epsilon}}

\newcommand{\etav}[0]{{\boldsymbol{\eta}}}

\newcommand{\thetav}{\boldsymbol{\theta}}

\newcommand{\given}{\,|\,}

\usepackage[symbol]{footmisc}

\renewcommand{\thefootnote}{\fnsymbol{footnote}}

\title{%Simple and Scalable 
Bayesian Attention Modules} 
\author{Xinjie Fan$^{*,1}$, Shujian Zhang$^{*,1}$, Bo Chen$^2$, and Mingyuan Zhou$^1$\\
$^1$The University of Texas at Austin and $^2$Xidian University
\\
\texttt{xfan@utexas.edu, szhang19@utexas.edu,}\\\texttt{bchen@mail.xidian.edu.cn, mingyuan.zhou@mccombs.utexas.edu} }

\begin{document}

\maketitle

\begin{abstract}
%\sz{or the title as "Simple and Scalable Bayesian Attention Neural Networks}
Attention modules, as simple and effective tools, have not only enabled deep neural networks to achieve state-of-the-art results in many domains, but also enhanced their interpretability. Most current models use deterministic attention modules due to their simplicity and ease of optimization. Stochastic counterparts, on the other hand, are less popular despite their potential benefits. The main reason is that stochastic attention often introduces optimization issues or requires significant model changes. In this paper, we propose a scalable stochastic version of attention that is easy to implement and optimize. We construct simplex-constrained attention distributions by normalizing reparameterizable distributions, making the training process differentiable. We learn their parameters in a Bayesian framework where a data-dependent prior is introduced for regularization. We apply the proposed stochastic attention modules to various attention-based models, with applications to graph node classification, visual question answering, image captioning, machine translation, and language understanding. Our experiments show the proposed method brings consistent improvements over the corresponding baselines.{\let\thefootnote\relax\footnote{{$^*$ Equal contribution.
%} \footnote[]{
Corresponding to: \texttt{mingyuan.zhou@mccombs.utexas.edu}}}} 
\end{abstract}

\section{Introduction}
Attention modules, aggregating features with weights obtained by aligning latent states, have become critical components for state-of-the-art neural network models in various applications, such as natural language processing \cite{bahdanau2014neural,vaswani2017attention,devlin2018bert,lan2019albert}, computer vision \cite{bello2019attention,ramachandran2019stand}, %,hu2018squeeze}, 
graph analysis \cite{velivckovic2017graph,lee2019attention}, and multi-modal learning \cite{xu2015show,rennie2017self,
yu2019deep,wang2019reinforced}. They have been proven to be effective in not only being combined with other types of neural network components, such as recurrent \cite{bahdanau2014neural,xu2015show} and convolutional units \cite{wang2018non,bello2019attention}, 
but also %serving as the sole type of components
being used to build a stand-alone architecture % that works by itself
\cite{vaswani2017attention,ramachandran2019stand,parmar2018image}. Besides boosting the performance, %attention modules
using them also often helps aid model visualization and enhance interpretability %interpretation. 
%bility are also common advantages of 
\cite{bahdanau2014neural,xu2015show}.

While the attention mechanism provides useful inductive bias, 
attention weights are often treated as 
%it is common for attention-based models to treat the  as 
deterministic rather than random variables. Consequently, %Often, if no dropout layer is included, 
the only source of randomness lies at %such models is in 
the model output layer. For example, in classification models, the randomness is in the final logistic regression layer, while in discrete sequence generation, it is in the conditional categorical output layer. % distributions. 
However, a single stochastic output layer is often insufficient in modeling %cannot model 
complex dependencies %efficiently 
\cite{chung2015recurrent}. The idea of augmenting deterministic neural networks with latent random variables has achieved success in many fields to model highly structured data, such as 
texts \cite{zhou2016augmentable,zhang2018whai},
%natural 
speeches \cite{chung2015recurrent,fraccaro2016sequential,bayer2014learning}, natural language sequences \cite{bowman2015generating,he2018sequence,zhang2016variational,guo2020recurrent}, and images \cite{kingma2013auto,higgins2017beta,blundell2015weight}. Such modification may not only boost the performance, but also provide better uncertainty estimation \cite{gal2016dropout,gal2017concrete}. 

As attention weights can be interpreted as alignment weights, it is intuitive to connect the attention module with latent alignment models \cite{xu2015show,deng2018latent,lawson2018learning}, where latent alignment variables are stochastic and the objective becomes a lower bound of the log marginal likelihood. Making the attention weights stochastic and learning the alignment distribution in a probabilistic manner brings several potential advantages. First, adding latent random variable enhances the model's ability to capture complicated dependencies in the target data distribution. Second, we are able to adopt Bayesian inference, where we may build our prior knowledge into prior regularization on the attention weights and utilize posterior inference to provide a better basis for model analysis and uncertainty estimation \cite{deng2018latent,gal2017concrete}. 

Most current work on stochastic attention focus on hard attention \cite{xu2015show,shankar2018posterior,deng2018latent}, where the attention weights are discrete random variables sampled from categorical distributions. However, standard backpropagation no longer applies to the training process of such models and one often resorts to a REINFORCE gradient estimator \cite{williams1992simple}, which has large variance.
%either a REINFORCE gradient estimator \cite{williams1992simple}, which is unbiased but has large variance, or a continuous relaxation based gradient estimator \cite{maddison2016concrete,jang2016categorical}, which has lower variance but is biased. 
%is often unstable due to the high variance of REINFORCE gradient estimators \cite{kingma2013auto,xu2015show,deng2018latent}
Such models generally underperform their deterministic counterparts, with a few exceptions, where a careful design of baselines and curriculum learning are required \cite{xu2015show,deng2018latent,lawson2018learning}. While attending to multiple positions at one time is intuitively more preferable, probabilistic soft attention is less explored.
\citeauthor{bahuleyan2017variational} \cite{bahuleyan2017variational}
propose to use the normal distribution to generate the attention weights, which are hence possibly negative and %attention weights 
do not sum to one. \citeauthor{deng2018latent} \cite{deng2018latent} consider sampling attention weights from the Dirichlet distribution, which is not reparameterizable and hence not amenable to gradient descent based optimization. 

In this paper, we propose Bayesian attention modules where the attention weights are treated as latent random variables, whose distribution parameters are obtained by aligning keys and queries. We satisfy the simplex constraint %($i.e.$, nonnegative and sum to one) 
on the attention weights, 
by normalizing the random variables drawn from either the Lognormal or Weibull distributions. Both distributions generate non-negative random numbers that are reparameterizable. 
%distributions over simplex by normalizing reparameterizable distributions,
%including .
In this way, the whole training process can be made differentiable via the reparameterization trick. 
%For regularization, 
We introduce a contextual prior distribution whose parameters are functions of keys to impose 
%from the optimization perspective can be viewed as 
a 
Kullback--Leibler (KL) divergence based regularization.
%term to the loss function. %a data reconstruction objective. 
To reduce the variance of gradient estimation, %estimating the KL term,
we pick the prior distribution such that the KL term can be rewritten 
in a semi-analytic form, $i.e.$, 
%as a summation of several terms, each of which is 
an expectation of analytic functions. % semi-analytic \rr{need to be more precise about "semi-analytica"}. 

Compared to previous stochastic attentions, our method is much simpler to implement, requires %surprisingly
only a few modifications to standard deterministic attention, is stable to train, and maintains good scalability, thereby making it attractive for large-scale deep learning applications. We evaluate the proposed stochastic attention module on a broad range of tasks, including graph node classification, visual question answering, image captioning, machine translation, and language understanding, where attention plays an important role. % and a variety of attention types are included, such as self attention, guided attention, and encoder-decoder attention.
We show that %with only a slight increase in memory and computation, 
the proposed method %achieves better performance over
consistently outperforms baseline attention modules and provides better uncertainty estimation. Further, we conduct a number of ablation studies to reason the effectiveness of the proposed model.

% \textbf{Contributions: }
% \begin{itemize}
% \item using weibull, and gamma, the KL term is step-wise analytic, variance is reduced further (implicit dirichlet prior).
% \item conduct experiments on classification task VQA and sequence generation task image captioning show improvement and visualization, sparsity. UNCERTAINTY
% \end{itemize}

% Attention is summarize the information using some weights. These weights are computed by?????

% In Bayesian perspective, attention can be viewed as updating the weights based on the query and key information.

\section{Preliminaries on attention modules}
% Attention modules has become increasingly popular not only 
% in combination with other types of neural network units such as recurrent \cite{xu2015show,bahdanau2014neural} and convolutional units \cite{wang2018non,bello2019attention}, but also as a standalone architecture that works by itself \cite{vaswani2017attention,ramachandran2019stand,parmar2018image}. 
In this section, we briefly review the standard deterministic soft attention modules
that have been widely used in various neural networks.

\textbf{Basic module:} Consider $n$ key-value pairs, packed into a key matrix $K\in \mathbb{R}^{n\times d_k}$ and a value matrix $V\in \mathbb{R}^{n\times d_v}$, and $m$
queries packed into $Q \in \mathbb{R}^{m\times d_k}$, where the dimensions of queries and keys are both equal to $d_k$. Depending on the applications, key, value, and query may have different meanings. For example, in self-attention layers \cite{vaswani2017attention}, key, value, and query are all from the same source, $i.e.$, the output of the previous layer and in this case $m$ equals to $n$. In encoder-decoder attention layers, the queries come from the decoder layer, while the keys and values come from the output of the encoder \cite{vaswani2017attention,xu2015show,bahdanau2014neural}. When attention is used for multi-modal cases, %the modality the queries come from differ from that the keys and values do.
the queries often come from one modality while the keys and values come from the other %, which is referred to as guided-attention 
\cite{yu2019deep}.
% \rr{explain In self attention, we have $m=n$. explain how K,Q,and V are related to the output of the previous layer; maybe consider moving the last paragraph of Section 2 to here or in the very beginning }

Attention modules make use of keys and queries to obtain \textit{deterministic} attention weights $W$, which are used to aggregate values $V$ into output features $O = W V \in \mathbb{R}^{m\times d_v}$. Specifically, $W$ is obtained through a softmax function across the key dimension as $W = \text{softmax}(f(Q, K)) \in \mathbb{R}^{m\times n}$, so that it is a non-negative matrix with each row summing to one. Thus if we denote $\Phi = f(Q,K)$, then 
\vspace{-1mm}
\ba{ W_{i,j} = \frac{\exp (\Phi_{i,j}) }{ \sum_{j'=1}^n \exp (\Phi_{i,j'})}.\label{eq:W}}
Intuitively, the scale of element $W_{i,j}$ represents the importance of the $j$th key to the $i$th query, and the neural network should learn $Q$ and $K$ such that $W$ gives higher weights to more important features. There are many choices of the alignment score function $f$, including scaled dot-product \cite{vaswani2017attention,devlin2018bert}, %where $f(Q,K)=\text{softmax}(QK^T/\sqrt{d_k})$, 
additive attention \cite{xu2015show,rennie2017self,bahdanau2014neural}, and several other variations \cite{luong2015effective,graves2014neural,wang2018non}. 

\textbf{Multi-head and multi-layer attention:} Multi-head attention is proposed to attend to information from different representation subspaces \cite{vaswani2017attention}, where queries, keys, and values are linearly projected $H$ times by $H$ different projection matrices, %linear projections, 
producing $H$ output values that are concatenated as the final attention layer output. One may then stack attention layers by placing one on top of another, leading to deep attention modules \cite{vaswani2017attention,yu2019deep}. For a deep attention module with $L$ attention layers and $H$ heads for each layer, the output of the $l$th layer would be $O^{l} = [W^{l,1}V^{l,1}, ..., W^{l,H}V^{l,H}],$ where $W^{l,h} = \text{softmax}(f(Q^{l,h}, K^{l,h}))$, $Q^{l,h} = Q^{l}M^{l,h}_Q$, $K^{l,h}=K^{l}M^{l,h}_K$, and $V^{l,h} = V^{l}M^{l,h}_V$ for $h=1,...,H$, and the $M$'s are parametric matrices that the neural network needs to learn. Then the output of the $l$th attention layer, $O^l$, is fed into the next attention layer (possibly after some transformations) and the queries, keys, and values of the $(l+1)$th layer would be functions of $O^l$.

%\xj{Any other attention types?}

\section{Bayesian attention modules: a general recipe for stochastic attention}
We suggest a general recipe for stochastic attention: 1) treat attention weights as data-dependent local random variables and learn their distributions in a Bayesian framework, 2) use normalized reparametrizable distributions to construct attention distributions over simplex, and 3) use a key-based contextual prior as regularization. 

% Below we show how we can treat attention weights as data-dependent local random variables and learn their distributions in a Bayesian framework. %Then, we show how to construct reparameterizable variational attention distributions and the corresponding contextual prior attention distributions. 

\subsection{Learning attention distributions in a Bayesian way}
Consider a supervised learning problem with training data $\mathcal{D}:=\{\xv_i, \yv_i\}_{i=1}^N$, where we model the conditional probability $p_{\thetav}(\yv_i\given\xv_i)$ using a neural network parameterized by~$\thetav$, %included in which are %the parameters of 
which includes
the attention projections $M$'s. 
For notational convenience, below we drop the data index $i$. Using vanilla attention modules, the mapping from $\xv$ to the likelihood $p_{\thetav}(\yv|\xv)$ is deterministic, so the whole model is differentiable meaning that it is tractable to directly maximize the likelihood. % is tractable. 

Now, we turn the mapping from queries and keys to attention weights $W$ stochastic. Instead of using deterministic weights obtained from queries and keys to aggregate values, we treat $W=\{W^{l,h}\}_{l=1:L, h=1:H}$ as a set of data-dependent local latent variables sampled from %distribution 
$q_{\phiv}$, which can be parameterized by some functions of queries and keys. Intuitively, we argue that this distribution can be viewed as a variational distribution approximating the posterior  of local attention weights $W$, under a Bayesian model, given the data $\xv, \yv$.  Therefore, we can learn $q_{\phiv}$ with amortized variational inference \cite{kingma2013auto}. 
Note that, unlike \citet{deng2018latent} and \citet{lawson2018learning}, we do not enforce  $q_{\phiv}$ to be dependent on $\yv$, which might not be available during testing. Instead, we use the queries and keys in standard attention modules to construct $q_{\phiv}$, so $q_{\phiv}$ %could be independent of $\yv$ and only 
depends on $\xv$ only or both $\xv$
and the part of $\yv$ that has already been observed or generated by the model.
%but not necessarily on $\yv$. 
For example, in visual question answering  or graph node classification, $q_{\phiv}$ only depends on input $\xv$. While in sequence generation, like image captioning or machine translation, $q_{\phiv}$ could be dependent on the observed part of $\yv$ as the queries come from~$\yv$. 

%\xj{I moved the remark here. }
Constructing variational distribution in such a way has several advantages. First, as we will show in the next section, by utilizing keys and values, transforming a set of deterministic attention weights into an attention distribution becomes straightforward and requires minimal changes to standard attention models. We can even easily adapt pretrained standard attention models for variational finetuning (shown in Section~\ref{sec:bert}). Otherwise, building an efficient variational distribution often requires domain knowledge \cite{deng2018latent} and case by case consideration. Second, due to a similar structure as standard attention modules, $q_{\phiv}$ introduces little additional memory %cost, or
and computational cost, for which we provide a complexity analysis in Section~\ref{sec:3.4}. Third, as keys and values are available for both training and testing, we can use the variational distribution $q_{\phiv}$ during testing. By contrast, previous works \cite{deng2018latent,lawson2018learning} enforce $q_{\phiv}$ to include information not available during testing,  %limits its use for training only. %
restricting its usage at the testing time. %for out-of-sample predictions. 
Further, as keys and queries depend on the realization of attention weights in previous layers, this structure naturally allows \textit{cross-layer dependency} between attention weights in different layers so that it is capable of modeling complex distributions.

% \xj{maybe remove this remark?? \mz{agree, could keep some of it as regular text} }
% \begin{remark}
% Consider the joint distribution $q_{\phiv}(W) = \prod_{l=1}^{L} q_{\phiv}(W_l|W_{1:l-1})$. The distribution $q_{\phiv}(W_l|W_{1:l-1})$ is parameterized by queries and keys in layer $l$, which are functions of the previous layers' attention weights, i.e., this parameterization of $q_{\phiv}$ allows \textbf{cross-layer dependency} between $W_l$ and $W_{1:l-1}$. %Therefore the joint distribution $q_{\phiv}(W) = \prod_{l=1}^{L} q_{\phiv}(W_l|W_{1:l-1})$ is not tractable \rr{I think what you really meant was the marginal is not tractable??} \xj{I meant the joint is not a standard family}, but
% And each conditional $q_{\phiv}(W_l |W_{1:l-1})$ is normalized independent Weibull or Lognormal distributions, due to \textbf{cross-head independence} in layer $l$ conditioned on previous layers. 
% \end{remark}

Consider a Bayesian model, where we have  prior $p_{\etav}(W)$  and likelihood $p_{\thetav}(\yv\given \xv, W)$ that share a common structure with vanilla deterministic soft attention. We learn the distribution $q_{\phiv}$ by minimizing $\mbox{KL}(q_{\phiv}(W)||p(W\given \xv,\yv))$, the KL divergence from the posterior distribution of $W$ given $\xv$ and $\yv$ to $q_{\phiv}$.
With amortized variational inference, 
it is equivalent to maximizing $\mathcal{L_{\mathcal D}}=\sum_{(\xv,\yv)\in \mathcal{D}} \mathcal{L}({\xv,\yv})$, an evidence lower bound (ELBO) \citep{hoffman2013stochastic,blei2017variational,deng2018latent}  of
the intractable log marginal likelihood $\sum_{(\xv,\yv)\in\mathcal{D}}\log p(\yv \given \xv)=\sum_{(\xv,\yv)\in\mathcal{D}} \log \int p_{\thetav}(\yv\given \xv, W)p_{\etav}(W)d W$,
%$\sum_{(\xv,\yv)\in \mathcal{D}}\ln \int p(\yv\given \xv,W)p_{\etav}(W)dW$ 
 where 
%\vspace{-3.5mm}
\ba{\small\textstyle \mathcal{L}({\xv,\yv}):=\E_{q_{\phiv}(W)}\left[ \log p_{\thetav}(\yv\given \xv, W)\right]-\mbox{KL}(q_{\phiv}(W)||p_\etav(W))=
\E_{q_{\phiv}(W)}\left[ \log \frac{p_{\thetav}(\yv \given \xv, W)p_\etav(W)}{q_{\phiv}(W)}\right].\nonumber}
% \rr{We may need to emphasize we are doing amortized variational inference with $p(W|K)$ and $q(W|Q,K)$}
%
%
Learning attention distribution $q_{\phiv}$ via %this
amortized variational inference %framework 
provides a natural regularization for $q_{\phiv}$ from  prior $p_{\etav}$, where we can inject our prior beliefs %knowledge 
on attention distributions. We will show  we can parameterize the prior distribution with keys, so that the prior distribution can be data-dependent and encode the importance information of each keys. 
% We argue that learning attention distribution $q_{\phiv}$ under this amortized variational inference framework has several advantages. First, it provides a natural regularization for $q_{\phiv}$ from the prior $p_{\etav}$, where we can inject our prior beliefs %knowledge 
% on attention distributions. We will show that we can parameterize the prior distribution with keys, so that the prior distribution can be data-dependent. 
% Essentially, this imposes soft attention sharing between different queries. 
% This prior distribution encodes the importance information of each keys. 
% And the posterior updates this information by further incorporating query information. 
Meanwhile, we can update $\thetav$ and $\etav$ to maximize the ELBO.
%, which corresponds to a lower bound of the intractable log marginal likelihood $\sum_{(\xv,\yv)\in\mathcal{D}}\log p(\yv \given \xv)=\sum_{(\xv,\yv)\in\mathcal{D}} \log \int p_{\thetav}(\yv\given \xv, W)p_{\etav}(W)d W$. 
As $q_{\phiv}$ becomes closer to the posterior, the ELBO becomes a tighter lower bound.

% Instead of using deterministic weights obtained from queries and keys to aggregate values,
% \citeauthor{xu2015show} \cite{xu2015show} and \citeauthor{deng2018latent} \cite{deng2018latent} propose to view $W$'s as latent random variables, sampled from a distribution $q_{\phiv}$ parameterized by some functions of queries and keys, which, we show in the following, can be viewed as approximating posterior distribution of attention weights after seeing the data. 

%
%
% In this case, to maximize the marginal $p(\yv\given \xv)$ that is often intractable, it is common to resort to variational inference \citep{hoffman2013stochastic,blei2017variational,deng2018latent} that instead optimizes an evidence lower bound (ELBO) as 
%
% \vspace{-3.5mm}
% $$
% \mathcal{L}:=\E_{q_{\phiv}(W)}\left[ \log p_{\thetav}(\yv\given \xv, W)\right]-\mbox{KL}(q_{\phiv}(W)||p_\etav(W))=
% \E_{q_{\phiv}(W)}\left[ \log \frac{p_{\thetav}(\yv \given \xv, W)p_\etav(W)}{q_{\phiv}(W)}\right].
% $$
%where $q_{\phiv}(W)$ is the variational distribution trying to approximate the posterior distribution $p(W|\yv, \xv).$ 
% In the framework of auto-encoding variational
% Bayes \cite{kingma2013auto}, $q_{\phiv}$ can be viewed as
% an encoder trying to approximate the posterior distribution which is proportional to the product of the decoder $p_\thetav$ and prior $p_\etav$, i.e., $p(W| \yv, \xv) \propto p_{\thetav}(\yv | \xv, W)p_\etav(W)$. 

\subsection{Reparameterizable attention distributions}\label{Sec:3.2}
A challenge of %learning 
using stochastic attention weights 
is to optimize their distribution parameters.
%models comes from the optimization of the ELBO. 
%Most current 
Existing 
methods \cite{xu2015show,shankar2018posterior,deng2018latent} construct attention distributions in a way that standard backpropagation based training no longer applies. Without carefully customizing a training procedure for each specific task, it is generally hard to learn such distributions. %In this section, 
Below we %avoid the optimization issue by introducing
introduce reparameterizable soft stochastic attentions that allow %us to % where we are able to 
%optimize 
%the ELBO %objective 
%to be effectively optimized 
effectively
optimizing the distribution parameters in a simple and general way. 

Our goal is to construct a reparameterizable attention distribution $q_{\phiv}$ %and prior $p_\etav$ %such that is a distribution 
over the simplex, $i.e.$, $W^{l,h}_{i,j}\geq 0$ and $\sum_j W^{l,h}_{i,j}=1$. While the Dirichlet distribution, satisfying the simplex constraint and encouraging sparsity, appears to be a natural choice, 
it is not reparameterizable and hence not amenable to gradient descent based optimization. Here, we consider satisfying the simplex 
constraint %constructing distributions over a simplex 
by normalizing random variables drawn from non-negative reparameterizable distributions. In particular, we consider the Weibull and Lognormal distributions. We choose them mainly because they both lead to optimization objectives that are simple to optimize, as described below.
%\sz{should we mention the reasons why we choose weibull and lognormal from reparameterizable distributions family? and why not other reparameterizable distributions?}

% %that . The Weibull distribution %is known as a distribution family close to 
% resembles the
% gamma distribution but is reparameterizable \cite{WHAI}. The Lognormal distribution is transformed from the normal distribution to take only non-negative real values. %that is commonly used as a reparameterizable family for real number domain. 
% We review important characteristics of two distributions as follows.
\textbf{Weibull distribution:} The Weibull distribution $S\sim\text{Weibull}(k,\lambda)$ has probability density function (PDF) $p(S\given k,\lambda) = \frac{k}{\lambda ^k}S^{k-1}e^{-(S/\lambda)^k}$, where $S\in \mathbb{R}_+$. Its expectation is $ \lambda\Gamma(1+1/k)$ and variance is $\lambda^2\big[\Gamma\left(1+{2}/{k}\right) - \left(\Gamma\left(1+1/k\right)\right)^2\big]$. %To sample from Weibull distribution, 
It is reparameterizable as % trick that
drawing $S\sim\text{Weibull}(k,\lambda)$ is equivalent to letting $S =\Tilde{g}(\epsilon):= \lambda(-\log(1-\epsilon))^{1/k},~ \epsilon\sim \text{Uniform}(0,1)$. %\textbf{Weibull and Gamma distribution: }
It resembles the gamma distribution, and with $\gamma$ denoted as the Euler–Mascheroni constant, 
%Denoting $\gamma$ as the Euler–Mascheroni constant, it is known that 
the KL divergence from the gamma to Weibull distributions has an analytic expression \cite{zhang2018whai} as \vspace{-1mm}
$$
\small\textstyle\mbox{KL}(\text{Weibull}(k,\lambda)||\text{Gamma}(\alpha, \beta)) = \frac{\gamma \alpha}{k}-\alpha \log \lambda +\log k + \beta\lambda \Gamma(1+\frac 1k) - \gamma - 1-\alpha \log \beta+\log \Gamma (\alpha).
\nonumber$$
\textbf{Lognormal distribution:} The Lognormal distribution $S\sim\text{Lognormal}(\mu,\sigma^2)$ has PDF $p(S\given \mu,\sigma) = \frac{1}{S\sigma \sqrt{2\pi}} \exp \left[-\frac{(\log S- \mu)^2}{2\sigma^2}\right]$, where $S\in \mathbb{R}_+$. Its expectation is $\exp (\mu +{{\sigma ^{2}}/{2}})$ and variance is $[\exp(\sigma ^{2})-1]\exp(2\mu +\sigma ^{2})$. It is also reparameterizable as drawing $S\sim\text{Lognormal}(\mu,\sigma^2)$ is equivalent to letting $S =\Tilde{g}(\epsilon)= \exp(\epsilon\sigma + \mu ), ~\epsilon\sim \mathcal{N}(0,1)$. %\textbf{Lognormal and Lognormal distribution: }Similarly, 
The KL divergence is analytic as %from the Lognormal to Lognormal has an analytic expression as
$$\textstyle\mbox{KL}(\text{Lognormal}(\mu_1,\sigma^2_1)||\text{Lognormal}(\mu_2, \sigma^2_2)) = \log \frac{\sigma_2}{\sigma_1} + \frac{\sigma_1^2+(\mu_1-\mu_2)^2}{2\sigma^2_2} - 0.5
.\nonumber$$
Sampling $S_{i,j}^{l,h}$ from either the Weibull or Lognormal distribution, we obtain the simplex-constrained random attention weights $W$ by applying a normalization function $\bar{g}$ over $S$ %, producing samples on a simplex:
as $W_{i}^{l,h}=\bar{g}(S_{i}^{l,h}) := S_{i}^{l,h} / \sum_j S_{i,j}^{l,h}$. Note $W$ is reparameterizable but often does not have an analytic PDF.
%while $W$ can be generated with reparameterization, its analytic PDF is often not available. 
%After the normalizing transformation, the distribution is still reparameterizable even without knowing the analytic PDF.

\textbf{Parameterizing variational attention distributions:} To change the deterministic mapping from the queries and keys to attention weights to a stochastic one, we use queries and keys to obtain the distribution parameters of unnormalized attention weights $S$, which further define the variational distribution of the normalized attention weights $W$.

With the \textit{Weibull} distribution, we treat $k$ as a global hyperparameter and let $\lambda^{l,h}_{i,j} = \exp (\Phi_{i,j}^{l,h}) / \Gamma(1 + 1/k)$, and like before $\Phi^{l,h} = f(Q^{l,h},K^{l,h})$. Then, we sample $S_{i,j}^{l,h}\sim\mbox{Weibull}(k, \lambda^{l,h}_{i,j})$, which is the same as letting 
$S_{i,j}^{l,h} = \exp(\Phi_{i,j}^{l,h})\frac{(-\log(1-\epsilon_{i,j}^{h,l}))^{1/k}}{\Gamma(1+1/k)},~\epsilon_{i,j}^{h,l}\sim\mbox{Uniform}(0,1)$
. With the \textit{Lognormal} distribution, we treat $\sigma$ as a global hyperparameter and let $\mu_{i,j}^{l,h} = \Phi_{i,j}^{l,h} - \sigma^2/2$. Then, we sample $S_{i,j}^{l,h}\sim\mbox{Lognormal}(\mu_{i,j}^{l,h},\sigma^2)$, which is the same as letting 
$
S_{i,j}^{l,h}=\exp(\Phi_{i,j}^{l,h})\exp (\epsilon_{i,j}^{h,l} \sigma- \sigma^2/2 ),~\epsilon_{i,j}^{h,l}\sim\mathcal{N}(0,1)
$.
%\begin{remark}
Note our parameterizations ensure that $\E[S_{i,j}^{l,h}]=\exp (\Phi_{i,j}^{l,h})$. Therefore, if, instead of sampling $S_{i,j}^{l,h}$ from either distribution, we use its expectation as a substitute, then the mapping becomes equivalent to that of vanilla soft attention, whose weights are defined as in \eqref{eq:W}.
In other words,
if we let $k$ of the Weibull distribution go to infinity, or $\sigma$ of the Lognormal distribution go to zero, which means the variance of $S_{i,j}^{l,h}$ goes to zero and the distribution becomes a point mass concentrated at the expectation, then the proposed stochastic soft attention reduces to deterministic soft attention. Therefore, the proposed stochastic soft attention can be viewed as a generalization of vanilla deterministic soft attention.
%\end{remark}

%since $W_{i,j}^{l,h}=S_{i,j}^{l,h} / \sum_j S_{i,j}^{l,h} = \exp(\Phi_{i,j}^{l,h}) / \sum _j \exp(\Phi_{i,j}^{l,h})$. %\end{remark}

We have now constructed %the variational distribution 
$q_{\phiv}$ to be a reparameterizable distribution $W =g_{\phiv}(\epsilonv):=\bar{g}(\Tilde{g}_{\phiv}(\epsilonv))$, where $\epsilonv$ is a collection of $i.i.d.$ random noises with the same size as $W$. 
%The ELBO 
% $\mathcal{L}:=\E_{q_{\phiv}(W)}\left[ \log \frac{p_{\thetav}(y\given \xv, W)p_\etav(W)}{q_{\phiv}(W)}\right]$ 
%can be rewritten as 
%\vspace{-3mm}
% \ba{\textstyle
% \mathcal{L}:=\sum_{(\xv,\yv)\in\mathcal D}%=\E_{q_{\phiv}(W)}\left[ \log \frac{p_{\thetav}(y\given \xv, W)p_\etav(W)}{q_{\phiv}(W)}\right]=
% \E_{\epsilonv}\left[ \log \frac{p_{\thetav}(y\given \xv, g_{\phiv}(\epsilonv))p_\etav(g_{\phiv}(\epsilonv))}{q_{\phiv}(g_{\phiv}(\epsilonv))}\right].
% =\E_{\epsilonv}\left[ \log p_{\thetav}(y\given \xv, g_{\phiv}(\epsilonv))\right] - \mbox{KL}(q_{\phiv}(W)||p_{\etav}(W)).
% }
%And the derivative of $\mathcal{L}$ w.r.t $\phi$ is :
%$$\frac{\partial \mathcal{L}}{\partial \phiv} = \E_{\epsilon}\left[\frac{\partial \log p_{\thetav}(y\given \xv, g_{\phiv}(\epsilon))p_\etav(g_{\phiv}(\epsilon))}{\partial \phiv} - \frac{\partial \log q_{\phiv}(g_{\phiv}(\epsilon))}{\partial \phiv}\right].$$
%
To estimate the gradient of the ELBO, % w.r.t $\phi$, we could just draw multiple samples of $\epsilon$ and use Monte Carlo estimates as gradient estimators\cite{kingma2013auto}. The gradients for $\theta, \etav$ are estimated in a similar way. However, to obtain gradient estimates, 
however, we need either the analytic forms of both $p_\etav$ and $q_{\phiv}$, or the analytic form of the KL term, neither of which % in the KL penalty part, 
%which
are available. In the next section, we show how to work around this issue by imposing the KL regularization for latent variable $S$ before normalization and decomposing the joint distribution into a sequence of conditionals.

\subsection{Contextual prior: key-dependent Bayesian regularization}
%We have discussed how to construct a reparameterizable variational distribution $q_{\phiv}$ for attention weights $W$ using the keys and queries. We update the parameters $\theta, \phi, \etav,$ of decoder, encoder, and prior networks respectively with the gradients of ELBO. 
In the ELBO objective, there is a built-in regularization term, $i.e.$, the KL divergence from the prior distribution $p_\etav$ to variational distribution $q_{\phiv}$. To estimate the gradients, we need to %be able to 
evaluate $q_{\phiv}(W)$ and $p_\etav(W)$ for given attention weights $W$. We note that even though the analytic form of $q_{\phiv}(W)$ is not available, $q_{\phiv}(S)=\prod_{l=1}^{L} q_{\phiv}(S_l\given S_{1:l-1})$ is a product of %multivariate 
analytic PDFs (Weibull or Lognormal) for unnormalized weights $S$, so we rewrite the ELBO in terms of $S$ (we keep using $q_{\phiv}, p_{\etav}$ for $S$ as the distribution of $W$ is defined by $S$),
\vspace{-1mm}
% \ba{ \textstyle %\small
% \mathcal{L}:=\sum_{(\xv,\yv)\in\mathcal D}\big\{\E_{q_{\phiv}(S)}\left[ \log p_{\thetav}(\yv\given \xv, S)\right]-\mbox{KL}(q_{\phiv}(S)||p_\etav(S))\big\}\label{eq:elbo}
% }
\ba{ \textstyle %\small
\mathcal{L}(\xv, \yv):=\E_{q_{\phiv}(S)}\left[ \log p_{\thetav}(\yv\given \xv, S)\right]-\mbox{KL}(q_{\phiv}(S)||p_\etav(S))\label{eq:elbo}
}
% Therefore, as long as we have analytic form for $p_\etav(S)$, we can estimate the gradients w.r.t. ELBO. 
% \vspace{-1mm}
Note the KL divergence, 
%from the gamma to Weibull and that from the Lognormal to Lognormal are analytic, 
as shown in Section
\ref{Sec:3.2}, can be made analytic and hence it is natural to %. However, even if we 
 use either the gamma or Lognormal distribution to construct $p_\etav(S)$. %\xj{do we wanna say we prefer gamma over lognormal?, as we have not done much sparsity analysis.}
% To encourage sparse attention weights, we would
% %We %suspect %that the gamma is preferred 
% prefer the gamma
% to Lognormal, as %explained below. % for the following reason.
% %\begin{remark}
% %As
% %normalized %same-shaped 
% %gamma random variables with the same scale parameter follow a Dirichlet distribution, 
% imposing a $\mbox{Gamma}(\alpha,\beta)$ prior on $S$ implies a Dirichlet prior for $W$ if $\beta$ is the same across all $S$'s; if we let $\alpha$ to be clearly smaller than $1$, the KL term would impose a sparsity regularization on the weights due to the property of the Dirichlet distribution.
% %\end{remark}
Regardless of whether the gamma or Lognormal is used to construct $p_\etav(S)$, due to 
%the hierarchical construction of $S$ that introduces 
the dependencies between different stochastic attention layers,
we do not have analytic expressions for $\mbox{KL}(q_{\phiv}(S)||p_\etav(S))$.
%, the KL terms inside the ELBO. 
Fortunately, as shown in Lemma~\ref{lem:rao_black}, by 
%For both Weibull and Lognormal, 
%with appropriate prior distributions $p_\etav$, 
decomposing the joint into a sequence of conditionals and
exploiting
these analytic KL divergence expressions, % can be combined with the
we can express each KL term in a semi-analytic form.  %The proof is deferred to Appendix~\ref{sec:proof_rao_black}. 
According to the Rao-Blackwellization theorem \cite{mcbook}, we can reduce the Monte Carlo estimation variance % variance of our Monte Carlo estimator 
by plugging in the analytic part.
\begin{lemma}\label{lem:rao_black}
The KL divergence from the prior to variational distributions is semi-analytic as\vspace{-1mm}
\ba{\emph{\mbox{KL}}(q_{\phiv}(S)||p_\etav(S)) = \sum\nolimits_{l=1}^L \E_{q_{\phiv}(S_{1:l-1})}\small \underbrace{\emph{\mbox{KL}}(q_{\phiv}(S_l|S_{1:l-1})||p_\etav(S_l|S_{1:l-1}))}_{\text{analytic}}\label{eq:rao_black}}\vspace{-7mm}
\end{lemma}
%\xj{
\begin{proof}
\begin{equation}
\begin{split}
 \mbox{KL}(q_{\phiv}(S)||p_\etav(S)) =& \E_{q_{\phiv}(S)}\left[ \sum_{l=1}^L(\log q_{\phiv}(S_l|S_{1:l-1}) - \log p_\etav(S_l|S_{1:l-1})) \right] \\
 =& \sum_{l=1}^L \E_{q_{\phiv}(S)}\left[ \log q_{\phiv}(S_l|S_{1:l-1}) - \log p_\etav(S_l|S_{1:l-1}) \right] \\
 =& \sum_{l=1}^L \E_{q_{\phiv}(S_{1:l-1})}\E_{q_{\phiv}(S_{l}|S_{1:l-1})}\left[ \log q_{\phiv}(S_l|S_{1:l-1}) - \log p_\etav(S_l|S_{1:l-1}) \right]. %\\
 %=& \sum_{l=1}^L \E_{q_{\phiv}(S_{1:l-1})} \underbrace{\mbox{KL}(q_{\phiv}(S_l|S_{1:l-1})||p_\etav(S_l|S_{1:l-1}))}_{\text{analytic}}. \\
\end{split}\label{eq:rao_black_app}
\end{equation}
\end{proof}

%}

 \begin{figure*}[t] \vspace{-5mm}
 \centering
 \includegraphics[width=0.9\textwidth,height=2.7cm]{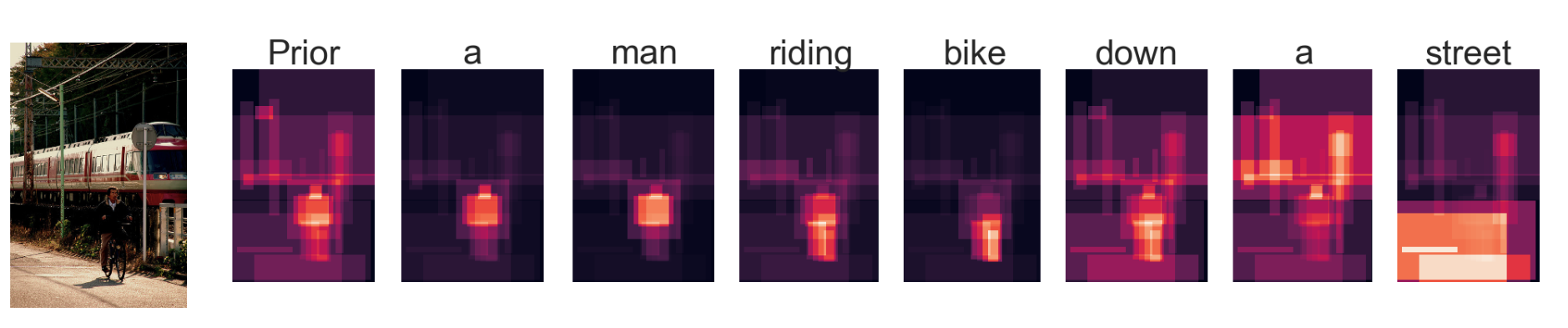}
 \vspace{-2.5mm}
 \caption{\small Visualization of attention weight samples from contextual prior distribution and variational distributions at each step for image captioning. Given the image, prior attention distribution over the image areas encodes the importance of each part before the caption generation process. Based on the prior distribution, the attention distribution can be updated at the each step using the current state of generation.} 
 \label{fig:att_vis_explanation} \vspace{-5mm}
\end{figure*}
\textbf{Key-based contextual prior: }Instead of treating the prior as a fixed distribution independent of the input $\xv$, here we make the prior depend on the input through keys. The motivation comes from our application in image captioning. 
% \sz{Visualization of attention weight. Given the image, prior attention distribution over the image areas encodes the importance of each part before the caption generation process. Based on the prior distribution, the attention distribution can be updated at the each step using the current state of generation.}
% The model includes \textbf{encoder-decoder attention} network where the keys comes from the images and the queries is based on the current state of generated sequences. 
Intuitively, given an image (keys), there should be a \textit{global} prior attention distribution over the image, indicating the importance of each part of the image even before the caption generation process. Based on the prior distribution, the attention distribution can be updated \textit{locally} using the current state of generation (queries) at the each step (see Figure~\ref{fig:att_vis_explanation}). This intuition can be extended to the general attention framework, where the prior distribution encodes the \textit{global} importance of each keys shared by all queries, while the posterior encodes the \textit{local} importance of each keys for each query. To obtain the prior parameters, we take a nonlinear transformation of the key features, followed by a softmax to obtain positive values and enable the interactions between keys. Formally, let $\Psi^{l,h} = \text{softmax}(F_2(F_{NL}(F_1(K^{l,h}))))\in \mathbb{R}^{n\times 1}$, where $F_1$ is linear mapping from $\mathbb{R}^{d_k}$ to a hidden dimension $\mathbb{R}^{d_\text{mid}}$, $F_2$ is linear mapping from $\mathbb{R}^{d_\text{mid}}$ to $\mathbb{R}$, and $F_{NL}$ denotes a nonlinear activation function, such as ReLU \cite{nair2010rectified}. With the gamma prior, % \textit{Gamma} distribution, 
we treat $\beta$ as a hyperparameter and let $\alpha^{l,h}_{i,j} = \Psi_{i,1}^{l,h}$. %\beta$. %(motivated by $\E_{x\sim \text{Gamma}(\alpha, \beta)}[x] = \alpha /\beta$). 
With the Lognormal, we treat $\sigma$ as a hyperparameter and let $\mu_{i,j}^{l,h} = \Psi_{i,1}^{l,h}$. % - \sigma^2/2$. 
%
% In visual question answering, the $\xv$ consists multimodal data. For \textbf{guided-attention}, the keys come from one modality, say images, and the queries come from the other, say questions. Before seeing the question, we should have a rough idea where the model should attend to in the images (keys, values). The prior belief can be updated using the information of questions (queries). For \textbf{self-attention}, the keys and queries come from one source. But prior attention distributions are shared across all tokens in the second layer, while posterior attention distributions vary.
%\textbf{KL divergence weight annealing: }
  Following previous work %on VAEs
\cite{bowman2016generating}, we add a weight $\lambda$
to the KL term and anneal it
from a small value to one.

\subsection{Putting it all together}\label{sec:3.4}
Combining \eqref{eq:elbo} and \eqref{eq:rao_black} and using reparameterization, we have $\mathcal{L}_{\lambda}(\xv,\yv)=\E_{\epsilonv}[\mathcal{L}_{\lambda}(\xv,\yv,\epsilonv)]$, where 
\vspace{-2mm}
\ba{
\mathcal{L}_{\lambda}(\xv,\yv,\epsilonv)=
%\sum_{(\xv,\yv)\in\mathcal D}
%\E_{\epsilonv}\bigg[
\log p_{\thetav}(\yv\given \xv, \Tilde{g}_{\phiv}(\epsilonv)) -{\lambda} \sum\nolimits_{l=1}^L\small \underbrace{\mbox{KL}(q_{\phiv}(S_l\given \Tilde{g}_{\phiv}(\epsilonv_{1:l-1}))||p_\etav(S_l\given \Tilde{g}_{\phiv}(\epsilonv_{1:l-1})))}_{\textit{analytic}}. %\bigg].
\label{eq:together}%\vspace{-2mm}
}
To estimate the gradient of $\mathcal{L}_\lambda(\xv,\yv)$ with respect to $\phiv, \thetav, \etav$, we compute the gradient of $\mathcal{L}_\lambda(\xv,\yv,\epsilonv)$, which is a Monte Carlo estimator with one sample of $\epsilonv$. This way provides unbiased and low-variance gradient estimates (see the pseudo code in Algorithm~\ref{alg:va} in Appendix).

% \textbf{Testing and uncertainty:}
At the testing stage, to obtain point estimates, we adopt the common practice of approximating the posterior means of prediction probabilities by 
substituting the latent variables by their posterior expectations \cite{srivastava2014dropout}. To calibrate estimation uncertainties, we draw multiple posterior samples, each of which produces one posterior prediction probability sample.

\textbf{Complexity analysis:} Our framework is computationally and memory efficient due to parameter sharing between the variational, prior, and likelihood networks. Extra memory cost comes from the contextual prior network which, for a single layer and single head attention, is of scale $O( d_kd_\text{mid})$. This is insignificant compared to the memory scale of $M$'s, $O(d_k d_v+ d_vd_v)$, as $d_\text{mid}$ is as small as $10\ll d_v$. Meanwhile, the additional computations involve the sampling process and computing the KL term which is of scale $O(m n)$. Computing the contextual prior is of scale $O(n d_k d_\text{mid})$. All above is inconsiderable compared to the computational scale of deterministic attentions, $O(m  n d_k d_v)$.
%Hence, the proposed method introduces little additional memory or computational cost.

% \xj{Add a discussion on the relation and difference with dropout?}
% \subsection{Compatibility with Reinforcement Learning}

% sparse attention can avoid mode collapse. (it can be viewed as an approximation to hard attention) (local attention). doubly attention.

% latent alignment assume independence for posterior. mean field assumption. additional neural network for posterior inference. additional cost for training. 

% Related papers:

% Estimating Uncertainty Online Against an Adversary

% Find the errors, get the better: Enhancing machine translation via word confidence estimation

% Can You Trust Your Model’s Uncertainty? Evaluating Predictive Uncertainty Under Dataset Shift

% Quantifying Uncertainties in Natural Language Processing Tasks

% Confidence Modeling for Neural Semantic Parsing

% Leveraging sentence similarity in natural language generation: Improving beam search using range voting

% relation between variational attention and dropout: log-normal is gaussian dropout applied to one layer. contextual gaussian dropout. 

% weibull: the beta in the gamma prior does not affect, it cancels out.

\section{Experiments}
Our method can be straightforwardly implemented in any attention based models. To test the general applicability of our method, we conduct experiments on a wide range of tasks where attention is essential, covering graphs (node classification),  multi-modal domains (visual question answering, image captioning), and natural language processing (machine translation, language understanding). A variety of attention types appear in these domains, including self, encoder-decoder, and guided attentions. In this section, we summarize the main experimental settings and results, and include the details in Appendix~\ref{sec:app_exp}. All experiments are conducted on a single Nvidia Tesla V100 GPU with 16 GB memory. Python code is available at \url{https://github.com/zhougroup/BAM}

% \rr{ could we make it becomes some type of layers from a library to call in Pytorch or TF?}

% \begin{table}[t]\centering
% \small
% \begin{sc}
% \caption{Model size comparison among different attentions. %\mz{Move this to later section after introducing MLP, WRN, and MCAN}
% }
% \label{table:parameter}\resizebox{0.65\columnwidth}{!}{
% \begin{tabular}{@{}l|lll@{}}\toprule
% Model & Soft & BAM+Fixed & BAM+Contextual \\ \midrule
% Image captioning & 267K & 58M \\ %(57812491)
% Visual question answering & 267K & 58M \\ %(57812491)
% \bottomrule
% \end{tabular}} %\vspace{-5mm}
% \end{sc}
% \end{table}

\subsection{Attention in graph neural networks}
We first adapt our method to graph attention networks (GAT)  \cite{velivckovic2017graph}, which leverages deterministic \textit{self-attention} layers to process node-features for graph node classification. The graph structure is encoded in the attention masks in a way that nodes can only attend to their neighborhoods' features in the graph. GAT %This attention-based framework 
is computationally efficient, capable of processing graphs of different sizes, and achieves state-of-the-art results on %across several 
benchmark graphs. We use the same model and experimental setup as in GAT \cite{velivckovic2017graph}, as summarized in Appendix~\ref{sec:app_graph}. We experiment with three %standard citation network 
benchmark graphs, including Cora, Citeseer, and Pubmed, %\cite{sen2008collective} 
for node classification in a transductive setting, meaning that training and testing are performed on different nodes of the same graph \cite{yang2016revisiting}. We include a summary of these datasets in Table~\ref{tab:gat_data} in Appendix. For large and sparse graph datasets like Pubmed, following GAT \cite{velivckovic2017graph}, we implement a sparse version of the proposed method, where sparse tensor operations are leveraged to limit the memory complexity to be linear in the number of edges.

\begin{table}[h] 
\vspace{-2mm}
\caption{Classification accuracy for graphs. }
\label{tab:gat_accuracy}
\centering
\resizebox{0.485\columnwidth}{!}{
\begin{tabular}{@{}llllll@{}}\toprule
Attention & Cora & Citeseer & PubMed\\ \midrule
GAT & 83.00 & 72.50 & 77.26 \\
BAM (NO KL) & 83.39 & 72.91 & 78.50 \\
BAM-LF  & 83.24 & 72.86 & 78.30\\
BAM-LC  & 83.34 & 73.04 & 78.76 \\
BAM-WF  & 83.48\small{$\pm0.2$} & 73.18\small{$\pm0.3$} & 78.50\small{$\pm0.3$}\\
BAM-WC & {\bf83.81}\small{$\pm0.3$} & {\bf73.52}\small{$\pm0.4$}& {\bf78.82}\small{$\pm0.3$} \\
\bottomrule
\end{tabular}}
\end{table}
{\bf Results.}
The results are summarized in Table~\ref{tab:gat_accuracy}. We report the results of soft attention (GAT), and $5$ 
% \begin{wraptable}[7]{r}{0.4\textwidth} 
% \vspace{-3mm}
% \caption{\small Classification accuracy for graphs. }
% \label{tab:gat_accuracy}
% \begin{sc}\vspace{-2mm}
% \resizebox{0.4\columnwidth}{!}{
% \begin{tabular}{@{}llllll@{}}\toprule
% Attention & Cora & Citeseer & PubMed\\ \midrule
% GAT & 83.00 & 72.50 & 77.26 \\
% BAM (NO KL) & 83.39 & 72.91 & 78.50 \\
% BAM-LF  & 83.24 & 72.86 & 78.30\\
% BAM-LC  & 83.34 & 73.04 & 78.76 \\
% BAM-WF  & 83.48\small{$\pm0.2$} & 73.18\small{$\pm0.3$} & 78.50\small{$\pm0.3$}\\
% BAM-WC & {\bf83.81}\small{$\pm0.3$} & {\bf73.52}\small{$\pm0.4$}& {\bf78.82}\small{$\pm0.3$} \\
% \bottomrule
% \end{tabular}}
% \end{sc} 
% \end{wraptable}
 versions of Bayesian Attention Modules (BAM): no KL regularization (NO KL), Lognormal with fixed prior (LF), Lognormal with contextual prior (LC), Weibull and fixed prior (WF), and Weibull and contextual prior (WC). We report the mean classification accuracies on test nodes over $5$ random runs, and the standard deviations of BAM-WC. { Note  the results of GAT are reproduced by running the code provided by the authors (\url{https://github.com/PetarV-/GAT}).}  Our results demonstrate that adapting the deterministic soft attention module in GAT to Bayesian attention %significantly 
 consistently improves the performance. Weibull distribution performs better that Lognormal, and contextual prior outperforms fixed prior and no prior. %The best results we achieve improve upon GAT by a margin of $0.81\%$, $1.02\%$, and $1.56\%$ on Cora, Citeseer, and Pubmed respectively. 

%\footnotetext{For Pubmed, we cannot achieve the GAT results reported in \citet{velivckovic2017graph}, possibly due to the instability of the sparse implementation.} 

%\mz{Can we say: For Pubmed, we achieve a worse result on GAT that the one reported in \citet{velivckovic2017graph}, possibly due to the discrepancy between the dense and sparse implementations of GAT; our model for Pubmed is built on top of the sparse implementation of GAT.} \xj{their result is also on sparse. do you think we should just remove the dataset. and say our algorithm can also be applied to sparse graph but it is instable so we do not include the result here?}

\subsection{Attention in visual question answering models}
We consider a multi-modal learning task, visual question answering (VQA), where a model predicts an answer to a question relevant to the content of a given image. The recently  proposed MCAN \cite{yu2019deep} uses \textit{self-attention} to learn the 
% VQA is a challenging task that requires 
fine-grained semantic meaning of both the image and  question, and \textit{guided-attention} to learn the reasoning between these two modalities. 
%to predict an accurate answer. Recently, \cite{yu2019deep} propose to use attention as building block  We adopted the MCAN \cite{yu2019deep} and the MCA (Modular Co-Attention) layers in MCAN are in composition of the self-attention(SA) units to model the dense intramodel interactions (word-to-word or region-to-region) and the guided-attention(GA) units to model the dense intermodal interactions (word-to-region). 
We apply BAM to MCAN and conduct experiments on the VQA-v2 dataset \cite{goyal2017making}, consisting of human-annotated question-answer   pairs for images from the MS-COCO dataset \cite{lin2014microsoft} (see detailed experiment settings in Appendix~\ref{sec:app_vqa}). 
%There are three types of questions: Yes/No, Number, and Other. There are 10 answers provided by 10 different human annotators for each question. 
To investigate the model’s robustness to noise, we also %construct a noisy version of the dataset
perturb the input by adding Gaussian noise to the image features \cite{larochelle2007empirical}. For evaluation, we consider both accuracy and uncertainty, which is necessary here as some questions are so challenging that even human annotators might have different answers. We use a hypothesis testing based Patch Accuracy vs Patch Uncertainty (PAvPU) \cite{mukhoti2018evaluating} to evaluate the quality of uncertainty estimation, which reflects whether the model is uncertain about its mistakes. We defer the details of this metric to Appendix~\ref{sec:uncer_pavpu}.

\begin{table}[h]
\vspace{-2mm} 
\centering
\caption{Results of different attentions on VQA.}%\vspace{-0.5mm} 
\label{tab:vqa_accuracy}
\begin{sc}
\resizebox{0.61\columnwidth}{!}{
\begin{tabular}{@{}lllcll@{}}\toprule
\small Metric & \multicolumn{2}{c}{\small Accuracy} & %\phantom{abc}
& \multicolumn{2}{c}{\small PAvPU}\\
\cmidrule{2-3} \cmidrule{5-6}
% & Original Data & Noisy Data && Original Data & Noisy Data \\ \midrule
\small Data & \small Original & \small Noisy && \small Original  & \small Noisy  \\ \midrule
%SENET & 0.0790 & NAN/NAN/NAN && 0.3670 & NAN/NAN/NAN \\
%MC dropout & 66.74 & 84.28 / 48.46 / 58.24 && 61.45 & 79.96 / 44.46 / 52.14 \\
\small Soft & 66.95\small & 61.25 && 70.04 & 65.34 \\
% Gaussian dropout & 70.55 & 80.73 / 60.26 / 65.52 && & \\
% Hard & 65.56 & & 60.21 && \\    do you know while this ta
\small BAM-LF & 66.89 & 61.43 && 69.92 & 65.48\\
\small BAM-LC & 66.93 & 61.58 && 70.14 & 65.60\\
\small BAM-WF & 66.93 & 61.60 && 70.09 & 65.62\\
% \small BAM-WC & 66.91 & {\bf62.89}\small && {\bf71.91}\small & \bf{66.75}\small\\
\small BAM-WC & {\bf67.02} \small{$\pm0.04$}& {\bf62.89} \small{$\pm0.06$} && {\bf71.21} \small{$\pm0.06$} & \bf{66.75} \small{$\pm0.08$}\\
\bottomrule
\end{tabular}}
\end{sc} 
\end{table}

\textbf{Results.} In Table~\ref{tab:vqa_accuracy}, we report the accuracy and uncertainty for both original and noisy data (see complete results in Table~\ref{tab:vqa_accuracy_complete}). In terms of accuracy, BAM performs similarly as soft attention on the original dataset, but clearly outperforms it %significantly 
on the more challenging noisy dataset, showing that stochastic soft attention is more robust to noise than deterministic ones. For uncertainty, as soft attention is deterministic we use the dropout in the model to obtain uncertainty, while for BAM we use both dropout and stochastic attention to obtain uncertainty. We observe that on both original and noisy datasets, BAM has better uncertainty estimations, meaning in general it is more uncertain on its mistakes and more certain on its correct predictions. {We provide qualitative analysis for uncertainty estimation by visualizing the predictions and uncertainties of three VQA examples in Figure \ref{fig:vqa_vis_explanation} in Appendix.} We note that we again observe the improvement from using contextual prior and that Weibull also performs better than Lognormal.

\subsection{Attention in image captioning models}

We further experiment a multi-modal sequence generation task, image captioning, %which is %one of \textit{the first} 
%a domain
where probabilistic attention (hard attention) was found to outperform deterministic ones \cite{xu2015show}. Image captioning models map an image $\xv$ to a sentence $\yv=(y_1,\ldots,y_T)$ that summarizes the image information. \textit{Encoder-decoder attention} is commonly adopted in state-of-the-art models. During encoding, bottom-up bounding box features \cite{anderson2018bottom} are extracted from images by a pretrained Faster R-CNN \cite{ren2015faster}. At each step of decoding, a weighted sum of bounding box features is injected into the hidden states of an LSTM-based RNN \cite{hochreiter1997long,mikolov2010recurrent} to generate words. The weights are computed by aligning the bounding box features (keys) and  hidden states from the last step (queries). %We use the attention structure, ``Att2in'', proposed by \citep{rennie2017self}, and replace the Resnet-encoded features by bounding box features. 
We conduct our experiments on MS-COCO \citep{lin2014microsoft}, following the setup of \citet{luo2018discriminability}.
%that consists of 123,287 images. Each image has at least five captions. We use the standard data split from \citet{karpathy2015deep}, with 113,287 training, 5000 validation, and 5000 testing images. %The vocabulary size $V$ is 9488 and the max caption length $T$ is 16.
For the model architecture, we employ an attention-based model (Att2in) of \citet{rennie2017self}, which we implement based on the code by \citet{luo2018discriminability} and replace the ResNet-encoded features by bounding box features (see details in Appendix~\ref{sec:app_ic}). In all experiments, we use maximum likelihood estimation (MLE) for training; we do not consider % alone without
reinforcement learning based fine-tuning \cite{rennie2017self,ranzato2015sequence,Fan2020Adaptive}, which is beyond the scope of this paper and we leave it as future work.  We report four widely used evaluation metrics, including
%\xj{using automatic evaluation metrics, 
BLEU \cite{papineni2002bleu}, CIDEr \cite{vedantam2015cider}, ROUGE \cite{lin-2004-rouge}, and METEOR \cite{banerjee-lavie-2005-meteor}. 
%}

%\textbf{Baselines.} 

% \begin{wraptable}[8]{r}{0.7\textwidth} 
\begin{table}[h]
\vspace{-2mm}
\caption{Comparing different attention modules on image captioning. } %\vspace{-3.5mm}
\label{tab:ic_main}
%\vskip 0.15in
\begin{center}
\begin{small}
\begin{sc}
\resizebox{0.9\columnwidth}{!}{
\begin{tabular}{@{}ccccccccc@{}}\toprule
Attention & BLEU-4 & BLEU-3 & BLEU-2 & BLEU-1 & CIDEr & ROUGE & METEOR\\ \midrule
Soft\cite{xu2015show} & 24.3 & 34.4 & 49.2 & 70.7 & - &- &23.9 \\
Hard\cite{xu2015show} & 25.0 & 35.7 & 50.4&71.8 & - &- &23.0 \\
Soft (Ours) & 32.2 & 43.6 & 58.3 & 74.9 & 104.0 & 54.7 & 26.1 \\
Hard (Ours) &26.5 & 37.2& 51.9& 69.8&84.4 &50.7&23.3 \\
% Soft+Gaussian & - & - & - & - & - & - & - \\
% BAM-LF & 32.4 & 43.7 & 58.3 & 74.9 & 104.2 & 54.7 & 26.1 \\
% BAM-WF & 32.7& 43.9&58.5 &75.0 &104.3 &54.8 &{\bf26.3}\\
BAM-LC &32.7 & 44.0 & 58.7 & 75.1 & {\bf 105.0}&54.8&{\bf26.3}\\
BAM-WC & {\bf 32.8}\small{$\pm0.1$} & {\bf 44.1}\small{$\pm0.1$}& {\bf 58.8}\small{$\pm0.1$}& {\bf 75.3}\small{$\pm0.1$}& { 104.5}\small{$\pm0.1$}& {\bf 54.9}\small{$\pm0.1$}&  26.2\small{$\pm0.1$}\\
\bottomrule
\end{tabular}}
\end{sc}
\end{small}
\end{center}
%\vskip -0.1in
\end{table}
% \end{wraptable}
\textbf{Results.} We incorporate the results of both deterministic soft attention and probabilistic hard attention from \citet{xu2015show}. We also report those results based on an improved network architecture used by BAM.  %Further, we compare Lognormal fixed prior results of BAM with fixed prior for both  and Weibull cases for comparison (BAM-LF, BAM-WF). 
%\textbf{Results.} 
Results in Table~\ref{tab:ic_main} show that the proposed probabilistic soft attention module (BAM) consistently outperforms the deterministic ones. In our implementation, we observe that it is difficult to make hard attention work well due to the high variance of gradients. Moreover, we experiment on modeling the attention weights as Gaussian distribution directly as in \citet{bahuleyan2017variational}. Our experiment shows that naively modeling attention weights with Gaussian distribution would easily lead to NAN results, as it allows the attention weights to be negative and not sum to $1$. 
% Therefore, we exclude the result from the table. 
Therefore, it is desirable %necessary 
to model attention weights with simplex constrained distributions.
%with distributions over simplex. 
We also experiment with BAM where the KL term is completely sampled and observe that the training becomes very unstable and often lead to NAN results. Therefore, constructing prior and posterior in a way that the KL term is semi-analytic does bring clear advantages to our model. In Figure~\ref{fig:att_vis_explanation}, we visualize the prior attention weights and posterior attention weights at each step of generation, where we can visually see how the posterior attention adapts from prior across steps. Further, we again observe that BAM-WC ourperforms BAM-LC and hence for the following tasks, we focus on using BAM-WC.  %we also observe the consistent improvement from using contextual prior over fixed prior, verifying our intuition that feeding the information of key to prior would enhance the model's capacity and achieve better performance. 

% \textbf{Ablation study.}  We also study the effect of several important hyperparameters, including $k$ in Weibull distribution, $\beta$ in Gamma distribution and $\sigma$'s in the prior/posterior distributions. We show the relationship between the model performance and these hyperparameters in table \xj{what?}. %Further, we experiment learning these parameters as functions of $\xv$ but this usually lead to training issues and result in NAN.

% \begin{table}[t]
% \caption{Comparing different attention module on VQA.}
% \label{tab:vqa_uncertain_complete}
% \vskip 0.15in
% \begin{center}
% \begin{small}
% \begin{sc}
% \resizebox{0.45\columnwidth}{!}{
% \begin{tabular}{@{}ccc@{}}\toprule
% & Accuracy & Uncertainty \\ \midrule
% Soft attention & 66.95 & 71.83 \\
% BAM fixed prior & 66.93 & 70.83 \\
% BAM contextual prior & 66.91& 72.13 \\
% Soft attention (noise) & 61.51 & 68.03 \\
% BAM fixed prior (noise) & 61.82 & 68.22\\
% BAM contextual prior (noise) & 62.89 & 67.74 \\
% \bottomrule
% \end{tabular}}
% \resizebox{0.45\columnwidth}{!}{
% \begin{tabular}{@{}ccc@{}}\toprule
% & Accuracy & Uncertainty \\ \midrule
% Soft attention & 66.95 & 71.83 \\
% BAM fixed prior & 66.93 & 70.83 \\
% BAM contextual prior & 66.91& 72.13 \\
% Soft attention (noise) & 61.51 & 68.03 \\
% BAM fixed prior (noise) & 61.82 & 68.22\\
% BAM contextual prior (noise) & 62.89 & 67.74 \\
% \bottomrule
% \end{tabular}}
% \end{sc}
% \end{small}
% \end{center}
% \vskip -0.1in
% \end{table}

\subsection{Attention in neural machine translation}
We also experiment with neural machine translation, where we compare with the %state-of-the-art
variational attention method proposed by \citet{deng2018latent}. 
% which takes in a source sentence and generates a sequence as the translation for the input. Attention is essential not only for encoding the input sequence features and but also for efficiently aggregating encoded representations during decoding stage. 
% We apply the proposed method here and compare it with the state-of-the-art variational attention \cite{deng2018latent}. 
We follow them to set the base model structure and experimental setting  (see in Appendix~\ref{sec:app_nmt}), and adapt their deterministic attention to BAM-WC. We compare the BLEU score \cite{papineni2002bleu} of BAM-WC and several variants of variational attention in \citet{deng2018latent}.
% See Appendix~\ref{sec:nmt} for details of the dataset, model structure and experimental settings.
% IWSLT dataset\cite{ranzato2015sequence} is used which is a relatively small but a standard benchmark for experimental NMT models. With the same Byte Pair Encoding vocabulary of 14k token \cite{sennrich2015neural}, we conduct the experiments following the process in \cite{edunov2017classical}. 

% \xj{I commented all details, pls move them to appendix and polish the sentences.}
% \begin{wraptable}[7]{r}{0.43\textwidth} 
% \vspace{-5mm}
% \caption{Results on IWSLT.}
% \label{tab:nmt_accuracy}
% \begin{sc} \vspace{-3.2mm}
% \resizebox{0.42\columnwidth}{!}{\centering
% \begin{tabular}{@{}llllll@{}}\toprule
% Model & BLEU \\ \midrule
% %SENET & 0.0790 & NAN/NAN/NAN && 0.3670 & NAN/NAN/NAN \\
% %MC dropout & 66.74 & 84.28 / 48.46 / 58.24 && 61.45 & 79.96 / 44.46 / 52.14 \\
% Soft Attention & 32.77\\
% % Gaussian dropout & 70.55 & 80.73 / 60.26 / 65.52 && & \\
% % Hard & 65.56 & 83.10 / 46.55 / 57.26 \\
% % BAM-LF & 66.89 & 84.46 / 49.11 / 58.24 \\
% % BAM-LC & 66.93 & 84.58 / 49.05 / 58.24\\
% % Marginal Likelihood & 33.29\\
% % Hard Attention + Enum & 31.40 \\
% % Hard Attention + Sample & 31.00 \\
% Variational Relaxed Attention &30.05\\
% Variational Attention + Enum & 33.68 \\
% Variational Attention + Sample & 33.30\\
% % BAM-LC (Ours) & {33.63} \xj{remove this?}\\
% BAM-WC (Ours) & {\bf33.81}\small{$\pm0.02$}\\
% % BAM-orange (Ours) & & {\bf73.96}\small{$\pm0.4$}& \\
% \bottomrule
% \end{tabular}}
% \end{sc} 
% % \end{table}
% \end{wraptable}
% Dataset?? summarize in appendix?
\textbf{Results.} As shown in Table~\ref{tab:nmt_accuracy}, BAM-WC outperforms deterministic soft attention significantly in BLEU score with same-level computational cost. Moreover, compared to all variants of variational attention \cite{deng2018latent}, BAM-WC achieves better performance with much less training cost, as BAM does not require the training of a completely separate variational network. In Table~\ref{tb:cost} in Appendix, we compare the run time and number of parameters of BAM and variational attention [26], where we show that BAM achieves better results while being more efficient in both time and memory. It is interesting to note that in \citet{deng2018latent} variational relaxed attention (probabilistic soft attention) underperforms variational hard attention, while BAM-WC, 
which is also probabilistic soft attention, can achieve better results. 
One of the main reason is that BAM-WC is reparameterizable and has stable gradients, 
while \citet{deng2018latent} use the Dirichlet distribution which is not reparameterizable so the gradient estimations still have high variances despite the use of a rejection based sampling and implicit differentiation \cite{jankowiak2018pathwise}. Also, we note that, compared to \citet{deng2018latent}, our method is much more  general because we do not need to construct the variational distribution on a case-by-case basis. 
% \sz{if the similar computational cost, we outperform than soft attention and varitational relaxed attention....}

% Further, we study $k$, \xj{fill} \sz{working}

\begin{table} 
\caption{Results on IWSLT.}
\centering
\label{tab:nmt_accuracy}
\resizebox{0.40\columnwidth}{!}{\centering
\begin{tabular}{@{}llllll@{}}\toprule
Model & BLEU \\ \midrule
%SENET & 0.0790 & NAN/NAN/NAN && 0.3670 & NAN/NAN/NAN \\
%MC dropout & 66.74 & 84.28 / 48.46 / 58.24 && 61.45 & 79.96 / 44.46 / 52.14 \\
Soft Attention & 32.77\\
% Gaussian dropout & 70.55 & 80.73 / 60.26 / 65.52 && & \\
% Hard & 65.56 & 83.10 / 46.55 / 57.26 \\
% BAM-LF & 66.89 & 84.46 / 49.11 / 58.24 \\
% BAM-LC & 66.93 & 84.58 / 49.05 / 58.24\\
% Marginal Likelihood & 33.29\\
% Hard Attention + Enum & 31.40 \\
% Hard Attention + Sample & 31.00 \\
Variational Relaxed Attention &30.05\\
Variational Attention + Enum & 33.68 \\
Variational Attention + Sample & 33.30\\
% BAM-LC (Ours) & {33.63} \xj{remove this?}\\
BAM-WC (Ours) & {\bf33.81}\small{$\pm0.02$}\\
% BAM-orange (Ours) & & {\bf73.96}\small{$\pm0.4$}& \\
\bottomrule
\end{tabular}}
% \end{table}
\end{table}

\subsection{Attention in pretrained language models}\label{sec:bert}
Finally, we adapt the proposed method to finetune deterministic \textit{self-attention} \cite{vaswani2017attention} based language models pretrained on large corpora. Our variational distribution parameters use the pretrained parameters from the deterministic models, and we randomly initialize the parameters for contextual prior. Then we finetune BAM %in the proposed variational framework 
for downstream tasks. We conduct experiments on $8$ benckmark datasets from General Language Understanding Evaluation (GLUE) \cite{wang2018glue} and two versions of Stanford Question Answering Datasets (SQuAD) \cite{rajpurkar2016squad,rajpurkar2018know}. 
We leverage the state-of-the-art pretrained model, ALBERT \cite{lan2019albert}, which is a memory-efficient version of BERT \cite{devlin2018bert} with parameter sharing and embedding factorization. Our implementation is based on Huggingface
PyTorch Transformer \cite{wolf2019transformers} and we use the base version of ALBERT following the same  setting  \cite{lan2019albert} (summarized in Appendix~\ref{sec:app_plm}). 

\begin{table}[htp!]\vspace{-2mm}
\caption{Performance of BAM on GLUE and SQuAD benchmarks.}\vspace{-1mm}
\label{tab:nlp}
\begin{center}
\begin{small}
\begin{sc}
\resizebox{0.97\columnwidth}{!}{
\begin{tabular}{@{}ccccccccccccc@{}}\toprule
 & MRPC & CoLA & RTE & MNLI & QNLI & QQP & SST & STS & SQuAD 1.1 & SQuAD 2.0 \\ \midrule
ALBERT-base & 86.5 & 54.5 & 75.8 & 85.1 & 90.9 & {\bf 90.8} & 92.4 & 90.3& 80.86/88.70& 78.80/82.07\\
ALBERT-base+BAM-WC & {\bf 88.5} & {\bf 55.8} & {\bf 76.2} & {\bf 85.6} & {\bf 91.5} & { 90.7} & {\bf 92.7} &{\bf 91.1} & {\bf 81.40/88.82} & {\bf 78.97/82.23}\\
\bottomrule
\end{tabular}}
\end{sc}
\end{small}
\end{center}
%\vspace{-3mm}
\end{table}

\textbf{Results. }In Table~\ref{tab:nlp}, we compare the results of ALBERT, which uses deterministic soft attention finetuned on each dataset with those finetuned with BAM-WC, resuming from the same checkpoints. We observe consistent improvements from using BAM-WC in both GLUE and SQuAD datasets even by only using BAM at the finetuning stage. We leave as future work using BAM at the pretrain stage.

\section{Conclusion}
We have proposed a simple and scalable Bayesian attention module (BAM) that achieves strong performance on a broad range of tasks but requires surprisingly few modifications to standard deterministic attention. The attention weights are obtained by normalizing reparameterizable distributions parameterized by functions of keys and queries. We learn the distributions in a Bayesian framework, introducing a key-dependent contextual prior such that the KL term used for  regularization is semi-analytic. Our experiments on a variety of tasks, including graph node classification, visual question answering, image captioning, and machine translation, show that BAM consistently outperforms corresponding baselines  and provides better uncertainty  estimation
%while maintaining the simplicity as deterministic attention 
at the expense of only slightly increased computational and memory cost. 
% Like deterministic attention, BAM is easy to implement and optimize, introducing little extra computation or memory. We demonstrate the general applicability of BAM
% Over a wide range of language and imaging tasks, we show that BAM outperforms deterministic attention significantly while maintaining its general implementation and simplicity with little extra computation and great uncertainty estimation.
% over a wide range of tasks where we show that BAM brings consistent improvements over the corresponding baselines and provides better uncertainty.
Further, on language understanding benchmarks, we show it is %even 
possible to finetune a pretrained deterministic attention with BAM and achieve better performance than finetuning with the original deterministic soft attention. With  extensive experiments and ablation studies, we demonstrate the effectiveness of each component of the proposed architecture, and show that BAM can serve as an efficient alternative to deterministic attention in the versatile tool box of attention modules. 

%\newpage
\section*{Broader Impact}
Attention modules have become critical components for state-of-the-art neural network models in various applications, including computer vision, natural language processing, graph analysis, and multi-modal tasks, to name a few. While we show improvements brought by our work on five representative tasks from a broad range of domains, our framework is general enough that it could be used to improve potentially any attention based models. Also, our framework solves two main issues of previously proposed probabilistic attentions that restrict their popularity, $i.e.$, optimization difficulty and complicated model design. We hope that our work will encourage the community to pay more attention to stochastic attention and study from a probabilistic perspective. 

Considering that attention models have been adopted in many machine learning systems, our work could have an important impact on those systems, such as self-driving \cite{kim2017interpretable}, healthcare \cite{choi2016retain}, and recommender systems \cite{tay2018multi}. However, there are potential risks of applying such systems in real-life scenario, because the data we encounter in real-life is biased and long-tailed, and also the discrepancy between training data and testing data might be large. Therefore, an undue trust in deep learning models, incautious usage or imprecise interpretation of model output by inexperienced practitioners might lead to unexpected false reaction in real-life and  unexpected consequences. However, we see opportunities that our work can help mitigate the risks with uncertainty estimation. Knowing when mistakes happen would enable us to know when to ask for human-aid if needed for real-life applications \cite{ovadia2019can}. %Therefore, our work has the potential to advance current machine learning systems in both accuracy and uncertainty estimation. 

% In this case, the proposed stochastic attention has one prominent advantage over deterministic ones, as it naturally provides uncertainty by modeling weights distributions rather than deterministic weights.

% We also need to note that a limitation of the approaches is that BAM is incorporated in the fine-tune stage. With all experiments and studies, incorporating BAM in the pretrained stage might promote uncertainty estimation and improve performance even further.

% A limitation of the approaches is that BAM is incorporated in the fine-tune stage current experiments that perform constrained-action selection is that they can be overly conservative when compared to methods that constrain stated distributions directly, especially with datasets collected from mixtures of policies.

% people can investigate more reparameterizable distributions.
% However, we need to note that, in real-life, estimating uncertainty is often more difficult than the experiment settings \cite{ovadia2019can}. The data we encounter becomes more biased, and long-tailed. And the discrepancy between training data and testing data might be even larger. Therefore, failure to use the uncertainty cautiously might lead to a false reaction in real-life. 

%\newpage

\section*{Acknowledgements}
X. Fan, S. Zhang, and M. Zhou
acknowledge the support of Grants IIS-1812699 and ECCS-1952193 from the U.S. National Science Foundation, 
the support of NVIDIA Corporation with the donation of the Titan Xp GPU used
for this research, 
%The authors acknowledge
and the Texas Advanced Computing Center (TACC) at The University of Texas at Austin for providing HPC resources that have contributed to the research results reported within this paper (URL: \url{http://www.tacc.utexas.edu}). B. Chen acknowledges the support of the Program for Young Thousand Talent by Chinese Central Government, the 111 Project (No. B18039), NSFC (61771361), Shanxi Innovation Team Project, and the Innovation Fund of Xidian University.
% This research was supported in part by Award IIS1812699 from the U.S. National Science Foundation. 
% The authors acknowledge the support of NVIDIA Corporation with the donation of the Titan Xp GPU used
% for this research.
% The authors acknowledge the Texas Advanced Computing Center (TACC) at The University of Texas at Austin for providing HPC resources that have contributed to the research results reported within this paper. URL: \url{http://www.tacc.utexas.edu}

%\small
\bibliographystyle{unsrtnat}
\bibliography{reference.bib,MZhou.bib}
\normalsize

\newpage
\appendix

\begin{center}
  \Large{\bf Bayesian Attention Modules: Appendix}  
\end{center}

% \section{Proof for Lemma~\ref{lem:rao_black}}\label{sec:proof_rao_black}
% \begin{proof}
% \begin{equation}
% \begin{split}
%  \mbox{KL}(q_{\phiv}(S)||p_\etav(S)) =& \E_{q_{\phiv}(S)}\left[ \sum_{l=1}^L(\log q_{\phiv}(S_l|S_{1:l-1}) - \log p_\etav(S_l|S_{1:l-1})) \right] \\
%  =& \sum_{l=1}^L \E_{q_{\phiv}(S)}\left[ \log q_{\phiv}(S_l|S_{1:l-1}) - \log p_\etav(S_l|S_{1:l-1}) \right] \\
%  =& \sum_{l=1}^L \E_{q_{\phiv}(S_{1:l-1})}\E_{q_{\phiv}(S_{l}|S_{1:l-1})}\left[ \log q_{\phiv}(S_l|S_{1:l-1}) - \log p_\etav(S_l|S_{1:l-1}) \right] \\
%  =& \sum_{l=1}^L \E_{q_{\phiv}(S_{1:l-1})} \underbrace{\mbox{KL}(q_{\phiv}(S_l|S_{1:l-1})||p_\etav(S_l|S_{1:l-1}))}_{\text{analytic}}. \\
% \end{split}\label{eq:rao_black_app}
% \end{equation}
% \end{proof}

\section{Algorithm }
\begin{algorithm}[h!]
 \caption{Bayesian Attention Modules}
 \label{alg:va}
\begin{algorithmic}\small
 \STATE $\thetav, \etav, \phiv \leftarrow$ Initialze parameters, $t\leftarrow 0$, $\rho\leftarrow$ anneal rate
 \REPEAT
 \STATE $\{\xv_i,\yv_i\}_{i=1}^M\leftarrow$ Random minibatch of M datapoints (drawn from full dataset)
 \STATE $\{\epsilonv_i\}_{i=1}^M\leftarrow$ Random samples
 \STATE $\lambda = \text{sigmoid}(t* \rho)$
 \STATE Compute gradients $\frac 1M \nabla_{\thetav, \etav, \phiv}\sum_i %\hat
 {\mathcal{L}}_{\lambda}(\xv_i, \yv_i, \epsilonv_i)$ according to Eq.  \eqref{eq:together}
 \STATE Update $\thetav, \etav, \phiv$ with gradients, $t \leftarrow t +1$
 \UNTIL{convergence}
 \STATE {\bf return:} $\thetav, \etav, \phiv$
\end{algorithmic}
\end{algorithm}

\section{Experiment details}\label{sec:app_exp}

\subsection{Graph neural networks}\label{sec:app_graph}
\subsubsection{Model descriptions}
As in \citet{velivckovic2017graph}, we apply a two-layer GAT model. We summarize the graph attention layer %\cite{velivckovic2017graph} 
here. Denote the input node features as $\hv=\{ \hv_1, ..., \hv_N\}$, where $N$ is the number of nodes. Then, the self-attention weights is defined as:

$$\alpha_{ij}^h=\frac{\exp (\text{LeakyReLU}(\av^h[\mathbf{W}^h \hv_i ||\mathbf{W}^h \hv_j]))}{\sum_{k\in \mathcal{N}_i}\exp (\text{LeakyReLU}(\av^h[\mathbf{W}^h \hv_i ||\mathbf{W}^h \hv_k]))},$$
where $\av^h, \mathbf{W}^h$ are neural network weights for head $h$, and $\mathcal{N}_i$ is the set of neighbor nodes for node $i$. $||$ denotes concatenation.

The output $\hv'=\{ \hv_1', ..., \hv_N'\}$ is computed as:
$$\hv_i' = ||_{h=1}^H \sigma \left(\sum\nolimits_{j\in \mathcal{N}_i}\alpha_{ij}^h \mathbf{W}^h \hv_j\right).$$

\subsubsection{Detailed experimental settings}
We follow the same architectural hyperparameters as in \citet{velivckovic2017graph}. The first layer consists of $H = 8$ attention heads computing $8$ features each, and the second layer has a single head attention following an exponential linear unit (ELU) \cite{clevert2015fast} nonlinearity. 
% The second layer has a single attention head that computes $C$ features (where $C$ is the number of classes),
Then softmax is applied to obtain probabilities. During training, we apply $L2$ regularization with $\lambda = 0.0005$. Furthermore, dropout \cite{srivastava2014dropout} with $p = 0.6$ is applied to both layers’ inputs, as well as to the normalized attention coefficients. Pubmed requires
slight changes for hyperparameter: the second layer has $H = 8$ attention heads, and the $L2$ regularization weight is $\lambda = 0.001$. Models are initialized using Glorot initialization \cite{glorot2010understanding} and trained with cross-entropy loss using the Adam SGD optimizer \cite{kingma2014adam} with
an initial learning rate of $0.01$ for Pubmed, and $0.005$ for all other datasets. In both cases we use
an early stopping strategy on both the cross-entropy loss and accuracy on the validation nodes, with a patience of $100$ epochs. Here, we summarize the hyperparameters for BAM, including anneal rate $\rho$ (as in Algorithm~\ref{alg:va}), $\sigma_1$ and $\sigma_2$ for prior Lognormal and posterior Lognormal respectively, $k$ for Weibull distribution, $\alpha, \beta$ for Gamma distribution, hidden dimension for contextual prior $d_\text{mid}$. On Pubmed, we use anneal rate $\rho=0.2$ for all methods. For BAM-LF, $\sigma_1=1\mathrm{E}6$, $\sigma_2=1\mathrm{E}{\minus2}$. For BAM-LC, $\sigma_1=1\mathrm{E}5$, $\sigma_2=1\mathrm{E}{\minus2}$, and $d_\text{mid}=5$. For BAM-WF, $k=10$, $\beta=1\mathrm{E}{\minus8}$, $\alpha=1\mathrm{E}{\minus4}$. For BAM-WC, $k=10$, $\beta=1\mathrm{E}{\minus4}$, and $d_\text{mid}=5$. On Cora, for BAM-LF, $\sigma_1=1\mathrm{E}{15}$, $\sigma_2=1\mathrm{E}{\minus6}$, and $\rho=0.2$. For BAM-LC, $\sigma_1=1\mathrm{E}{15}$, $\sigma_2=1\mathrm{E}{\minus15}$, $\rho=0.1$, and $d_\text{mid}=1$. For BAM-WF, $k=1$, $\beta=1\mathrm{E}{\minus10}$, $\alpha=1\mathrm{E}{\minus15}$, and $\rho=0.2$. For BAM-WC, $k=1$, $\beta=1\mathrm{E}{\minus10}$, $\rho=0.1$, and $d_\text{mid}=1$. On Citeseer, we use anneal rate $0.1$ for all methods. for BAM-LF, $\sigma_1=1\mathrm{E}{15}$ and $\sigma_2=1\mathrm{E}{\minus6}$. For BAM-LC, $\sigma_1=1\mathrm{E}{15}$, $\sigma_2=1\mathrm{E}{\minus5}$, and $d_\text{mid}=1$. For BAM-WF, $k=100$, $\beta=1\mathrm{E}{\minus15}$, and $\alpha=1\mathrm{E}{\minus7}$. For BAM-WC, $k=100$, $\beta=1\mathrm{E}{\minus15}$, and $d_\text{mid}=1$.
\begin{table}\centering
\caption{Basic statistics on datasets for node classification on graphs.}
\label{tab:gat_data}
\begin{sc}
\resizebox{0.49\columnwidth}{!}{
\begin{tabular}{@{}llllll@{}}\toprule
 & Cora & Citeseer & PubMed \\ \midrule
\#Nodes & 2708 & 3327 & 19717 \\
\#Edges & 5429 & 4732 & 44338 \\
\#Features/Node & 1433 & 3703 & 500 \\
\#Classes & 7 & 6 & 3 \\
\#Training Nodes & 140 & 120 & 60 \\
\#Validation Nodes & 500 & 500 & 500 \\
\#Test Nodes & 1000 & 1000 & 1000 \\
\bottomrule
\end{tabular}}
\end{sc} 
\end{table}

\subsection{Visual question answering}\label{sec:app_vqa}
\subsubsection{Uncertainty evaluation via PAvPU}
\label{sec:uncer_pavpu}
% {\bf Uncertainty evaluation via PAvPU:} 
%Due to there are $10$ answers provided by $10$ different human annotators for each question, a good uncertainty estimation becomes even more necessary. 
% Many metrics have been proposed to evaluate the quality of uncertainty estimation such as calibrated probability estimates to measure model \cite{guo2017calibration,naeini2015obtaining,kuleshov2018accurate} and the entropy or mutual information between the predictive distribution and posterior \cite{mukhoti2018evaluating}. 
We adopt hypothesis testing to quantify the uncertainty of a model's prediction. 
%The output p-value is more interpretable, conveying the clear information about how confident the model is with its prediction. 
Consider $M$ posterior samples of predictive probabilities $\{\pv_m\}_{m=1}^M$, where $\pv_m$ is a vector with the same dimension as the %a $C$-dimensional vector, and $C$ is the 
number of classes. 
% For single-label classification models, $\pv_m$ is produced by a softmax layer and sums to one, while for multi-label classification models, $\pv_m$ is produced by a sigmoid layer and each element is between $0$~and~$1$. The former output is used in most image classification models, while the latter is often used in VQA where multiple answers could be true for a single input. In both cases, %to obtain a point prediction, we average over $\{\pv_m\}_{m=1}^M$ and predict the class with the highest probability. Now,
To quantify how confident our model is about its prediction, we evaluate whether the difference between the probabilities of the first and second highest classes (in terms of posterior means) % and second highest class 
is statistically significant with two-sample $t$-test. 
With the output $p$-values and a given threshold, we can determine whether a model is certain about its prediction. Then, we evaluate the uncertain using the Patch Accuracy vs Patch Uncertainty metric \cite{mukhoti2018evaluating} which is defined as % follows:
%\begin{equation}
 $\mathrm{PAvPU}={\left(n_{a c}+n_{i u}\right)}/{\left(n_{a c}+n_{a u}+n_{i c}+n_{i u}\right)}$, 
 %\nonumber%\vspace{-2mm}
%\end{equation}
where $n_{ac}, n_{au}, n_{ic}, n_{iu}$ are the numbers of accurate and certain, accurate and uncertain, inaccurate and certain, inaccurate and uncertain samples, respectively. Since for VQA, each sample has multiple annotations, the accuracy for each answer can be a number between $0$ and $1$ and it is defined as $\text{Acc}(ans) = \min \{{(\# \text{human that said } ans)}/{3},1\}.$ Then we generalize the PAvPU for VQA task accordingly:
% \bas{
% \resizebox{1\hsize}{!}{$n_{ac} = \sum_{ans: certain} \text{Acc}(ans),~n_{iu} = \sum_{ans: uncertain} 1-\text{Acc}(ans)$\,}\notag\\
% \resizebox{1\hsize}{!}{$n_{au} = \sum_{ans: uncertain} \text{Acc}(ans),~n_{ic} = \sum_{ans: certain} 1 -\text{Acc}(ans)$.}
% }
% \bas{
% \resizebox{0.45\hsize}{!}{$n_{ac} = \sum_{i} \text{Acc}_i \text{Cer}_i,~n_{iu} = \sum_{i} (1-\text{Acc}_i)(1- \text{Cer}_i)$\,},\notag
% \resizebox{0.45\hsize}{!}{$n_{au} = \sum_{i} \text{Acc}_i (1-\text{Cer}_i),~n_{ic} = \sum_{i} (1-\text{Acc}_i)(\text{Cer}_i)$\,},
% }
\bas{
n_{ac} &= \sum_{i} \text{Acc}_i \text{Cer}_i,~n_{iu} = \sum_{i} (1-\text{Acc}_i)(1- \text{Cer}_i),\\
n_{au} &= \sum_{i} \text{Acc}_i (1-\text{Cer}_i),~n_{ic} = \sum_{i} (1-\text{Acc}_i)(\text{Cer}_i),
}
where for the $i$th prediction $\text{Acc}_i$ is the accuracy and $\text{Cer}_i \in\{0,1\}$ is the certainty indicator. 

\subsubsection{Model descriptions}
We use the state-of-the-art VQA model, MCAN \cite{yu2019deep}, to conduct experiments. The basic component of MCAN is Modular Co-Attention (MCA) layer. The MCA layer is a modular composition of two basic attention units: the self-attention (SA) unit and the guided-attention (GA) unit, where the SA unit focuses on intra-modal interactions and GA unit focuses on inter-modal interactions. Both units follow the multi-head structure as in \citet{vaswani2017attention}, including the residual and layer normalization components. The only difference is that in GA, the queries come from a different modality (images) than the keys and values (questions). By stacking MCA layers, MCAN enables deep interactions between the question and image features. We adopt the encoder-decoder structure in MCAN \cite{yu2019deep} with six co-attention layers.

% SA unit is to model the dense intra-model interaction between each question word pair and SA-GA is to model the intra-model interaction between each image region pairs. By utilizing the two deep co-attention models, namely stacking and encoder-decoder , which consists of multiple MCA layers to gradually refine the attended image and question features. 

% \xj{add descriptions of MCAN model.}

\subsubsection{Detailed experimental settings}
We conduct experiments on the VQA-v2 dataset, which is split into the training (80k images and 444k QA pairs), validation (40k images and 214k QA pairs), and testing (80k images and 448k QA pairs) sets. The evaluation is conducted on the validation set as the true labels for the test set are not publicly available
\cite{deng2018latent}, which we need for uncertainty evaluation. For the noisy dataset, we add Gaussian noise (mean $0$, variance $5$) to image features. We follow the hyperparameters and other settings from \citet{yu2019deep}: the dimensionality of input image features, input question features, and fused multi-modal features are set to be $2048$, $512$, and $1024$, respectively. The latent dimensionality in the multi-head attention is $512$, the number of heads is set to $8$, and the latent dimensionality for each head is $64$. The dropout rate is $0.1$. The size of the answer vocabulary is set to $N = 3129$ using the strategy in \citet{teney2018tips}. To train the MCAN model, we use the Adam optimizer \cite{kingma2014adam} with $\beta_1 = 0.9$ and $\beta_2 = 0.98$. The learning rate is set to $\min(2.5t\mathrm{E}{\minus 5}, 1\mathrm{E}{\minus 4})$, where $t$ is the current epoch number starting from $1$. After $10$ epochs, the learning rate is decayed by $1/5$ every $2$ epochs. All the models are trained up to $13$ epochs with the same batch size~of~$64$. To tune the hyperparameters in BAM, we randomly hold out $20\%$ of the training set for validation. After tuning, we train on the whole training set and evaluate on the validation set. For BAM-LF, $\sigma_1=1\mathrm{E}{9}$, $\sigma_2=1\mathrm{E}{\minus9}$, and $\rho=0.2$. For BAM-LC, $\sigma_1=1\mathrm{E}{9}$, $\sigma_2=1\mathrm{E}{\minus9}$, $\rho=0.2$, and $d_\text{mid}=20$. For BAM-WF, $k=1000$, $\beta=1\mathrm{E}{\minus2}$, $\alpha=1\mathrm{E}{\minus3}$, and $\rho=0.2$. For BAM-WC, $k=1000$, $\beta=1\mathrm{E}{\minus6}$, $\rho=0.1$, and $d_\text{mid}=20$.

\subsubsection{More results}
\begin{figure}[h]
\centering
%\vspace{-4.5mm}
% \centering
\includegraphics[height=6.cm]{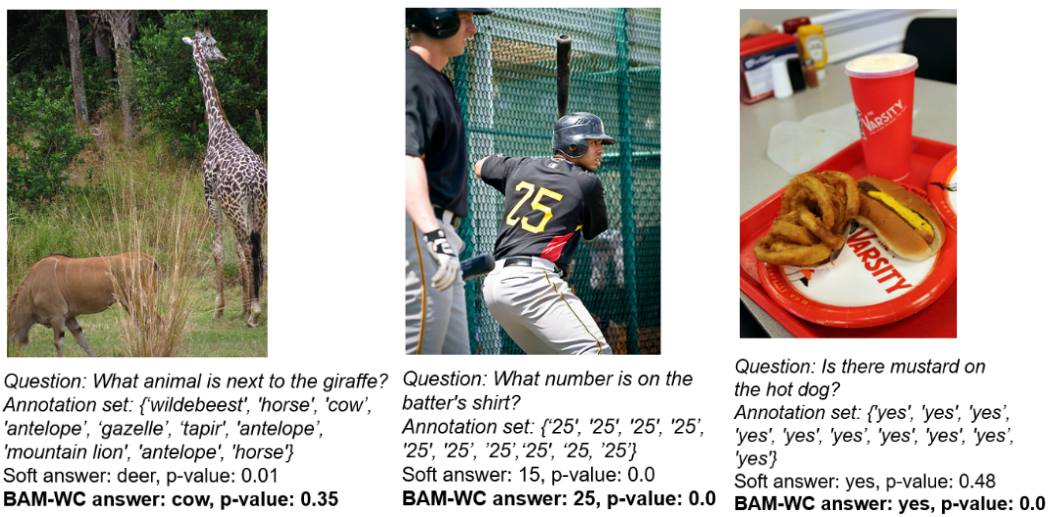}
\caption{\small VQA visualization: we present three image-question pairs along with human annotations. We show the predictions and uncertainty estimates of different methods. We evaluate methods based on their answers and $p$-values and highlight the better answer in bold (most preferred to least preferred: correct  certain > correct uncertain > incorrect uncertain > incorrect certain).}
\label{fig:vqa_vis_explanation} \vspace{-1.5mm}
\end{figure} 

\begin{table}[hbt!]\centering
\caption{Performance comparison of different attention modules on visual question answering.}
\label{tab:vqa_accuracy_complete}
%\begin{sc}
\resizebox{0.6\columnwidth}{!}{
\begin{tabular}{@{}lllcll@{}}\toprule
\multicolumn{6}{c}{Accuracy}\\\midrule
Attention & \multicolumn{2}{c}{Original Data} & %\phantom{abc}
& \multicolumn{2}{c}{Noisy Data}\\
\cmidrule{2-3} \cmidrule{5-6}
& ALL & Y/N / NUM / OTHER && ALL & Y/N / NUM / OTHER \\ \midrule
%SENET & 0.0790 & NAN/NAN/NAN && 0.3670 & NAN/NAN/NAN \\
%MC dropout & 66.74 & 84.28 / 48.46 / 58.24 && 61.45 & 79.96 / 44.46 / 52.14 \\
Soft & 66.95\small & 84.55 / 48.92 / 58.33 && 61.25 & 80.58 / 40.80 / 51.97 \\
% Gaussian dropout & 70.55 & 80.73 / 60.26 / 65.52 && & \\
% Hard & 65.56 & 83.10 / 46.55 / 57.26 && 60.21 & 79.39 / 38.76 / 51.31 \\
BAM-LF & 66.89 & 84.46 / 49.11 / 58.24 && 61.43 & 80.95 / 41.51 / 51.85 \\
BAM-LC & 66.93 & 84.58 / 49.05 / 58.24 && 61.58 & 80.70 / 41.31 / 52.40 \\
BAM-WF & 66.93 & 84.55 / 48.84 / 58.32 && 61.60 & 81.02 / 41.84 / 52.05 \\
% BAM-WC & 66.91 & 84.43 / 48.71 / 58.40 && {\bf62.89}\small & 81.94 / 41.90 / 53.96 \\
BAM-WC & {\bf67.02} & 84.66 / 48.88 / 58.42 && {\bf62.89}\small & 81.94 / 41.90 / 53.96 \\
\bottomrule
\end{tabular}}
%\end{sc} 
%\begin{sc}
\resizebox{0.6\columnwidth}{!}{
\begin{tabular}{@{}lllcll@{}}\toprule
\multicolumn{6}{c}{Uncertainty}\\\midrule
Attention & \multicolumn{2}{c}{Original Data} & %\phantom{abc}
& \multicolumn{2}{c}{Noisy Data}\\
\cmidrule{2-3} \cmidrule{5-6}
& ALL & Y/N / NUM / OTHER && ALL & Y/N / NUM / OTHER \\ \midrule
%SENET & 0.0790 & NAN/NAN/NAN && 0.3670 & NAN/NAN/NAN \\
Soft & 70.04 & 83.02 / 56.81 / 63.66 && 65.34 & 78.84 / 49.80 / 59.18 \\
% Gaussian dropout & 70.55 & 80.73 / 60.26 / 65.52 && & \\
% Hard & 68.98 & 81.29 / 56.29 / 62.98 && 64.84 & 76.89 / 50.56 / 59.45 \\
BAM-LF & 69.92 & 82.79 / 56.87 / 63.58 && 65.48 & 79.13 / 50.19 / 59.16 \\
BAM-LC & 70.14& 83.02 / 57.40 / 63.71 && 65.60 & 78.85 / 49.87 / 59.70 \\
BAM-WF & 70.09 & 83.00 / 56.85 / 63.78 && 65.62 & 79.16 / 50.22 / 59.40 \\
% BAM-WC & {\bf71.91}\small & 80.10 / 62.97 / 68.04 && \bf{66.75}\small & 80.21 / 51.38 / 60.58 \\
BAM-WC & {\bf71.21}\small & 83.95 / 58.12 / 63.82 && \bf{66.75}\small & 80.21 / 51.38 / 60.58 \\
\bottomrule
\end{tabular}}
%\end{sc} 
\end{table}

\subsection{Image captioning}\label{sec:app_ic}
\subsubsection{Model descriptions}
We conduct experiments on an attention-based model for image captioning, Att2in, in \citet{rennie2017self}. This model uses RNN as its decoder, and at each step of decoding, image features are aggregated using attention weights computed by aligning RNN states with the image features. Formally, suppose $I_1, ..., I_N$ are image features, $\hv_{t-1}$ is the hidden state of RNN at step $t-1$. Then, the attention weights at step $t$ are computed by:
$\alphav_t= \text{softmax} (\av_t+b_\alpha)$, and $a_t^i = W\text{tanh}(W_{aI}I_i+ W_{ah}\hv_{t-1}+b_a)$, where $W, W_{aI}, W_{ah}, b_\alpha, b_a$ are all neural network weights. Aggregated image feature $I_t= \sum_{i=1}^N \alpha_t^i I_i$ would then be injected into the computation of the next hidden state of RNN $\hv_t$ (see details in \citet{rennie2017self}).

\subsubsection{Detailed experimental settings}
We use the code from \url{https://github.com/ruotianluo/self-critical.pytorch} and conduct our experiments on the MS COCO dataset \citep{lin2014microsoft} that consists of 123,287 images. Each image has at least five captions. We use the standard data split from \citet{karpathy2015deep}, with 113,287 training, 5000 validation, and 5000 testing images. The vocabulary size $V$ is 9488 and the max caption length $T$ is 16. We replace the ResNet-encoded features in \citet{rennie2017self} with bounding box features extracted from a pre-trained Faster-RCNN \cite{ren2015faster} as visual features. Following the original setting in the code base, we use batch size $10$, Adam optimizer with learning rate $5\mathrm{E}\minus 4$, dropout rate of $0.5$ and train $30$ epochs. During training, we use MLE loss only without scheduled sampling or RL loss. For testing, we use greedy search to generate sequences. For BAM, we use contextual prior with $d_\text{mid}=10$ and $\rho=1$. For BAM-WC, $k=10$, $\beta=1\mathrm{E}\minus 6$. For BAM-LC, $\sigma_1=1\mathrm{E}3, \sigma_2=0.1$.

% Soft-weibull, contextual prior + se(hid size 10:, nonlinear=relu), k=10, kl_start_epoch=1, kl_anneal_rate=1, att_kl=1, alpha_gamma=contextual, beta_gamma=1e-6, index3:

% Soft-normal, contextual prior + se(hid size 10:, nonlinear=relu), sigma_prior=1e3, kl_start_epoch=1, kl_anneal_rate=1, att_kl=1, mu_prior=contextual, sigma_posterior=0.1, index2:

\subsection{Neural Machine Translation}\label{sec:app_nmt}
\subsubsection{Model descriptions}
To make comparision with \citet{deng2018latent}, we adopt the LSTM-based machine translation model in that paper.
%\cite{deng2018latent}. 
The model uses a bidirectional LSTM to encode a source sentence to source representations $\xv_1, ..., \xv_T$. At the step $j$ of decoding, current LSTM state $\tilde{\xv}$ (a function of previous target words $y_{1:j-1}$) is used as query. The attention weights is computed from an MLP between the query and encoded source token representations. Then the aggregated feature is used to produce the distribution over the next target work $y_j$ (see details in \citet{deng2018latent} or see code in \url{https://github.com/harvardnlp/var-attn}).

\subsubsection{Detailed experimental settings}
% \subsubsection{Dataset and Structure:}
\label{sec:nmt}

We use the same dataset, IWSLT \cite{cettolo2014report}, as \citet{deng2018latent}. We preprocess the data in the same way: using Byte Pair Encoding over the combined source/target training set to obtain a vocabulary size of 14,000 tokens \cite{sennrich2015neural}. We train on the sequence length up to 125. We use a two-layer bi-dreictional LSTM with 512 units and 768 units for the encoder and decoder, respectively. In addition, the batch size is $6$, dropout rate is $0.3$, learning rate is $3\mathrm{E}\minus4$ (Adam optimizer). %, parameter initialization over a uniform distribution $\mathcal{U}[-0.1,0.1]$
For testing, we use beam search with beam size 10 and length penalty $1$ \cite{wu2016google}. For BAM-WC, $k=5$, $\beta=1\mathrm{E}{\minus6}$, $\rho=1$, and $d_\text{mid}=5$. 
% We also recognize that the  $\beta$ has no impact on the result as the beta has been cancel out in the KL term. 
% In table \ref{tab:nmt_accuracy}, all methods including BAM are evaluated on BLEU (higher is better). Specifically, we perform the beam search with beam size 10 and length penalty $\alpha =1$ \cite{wu2016google} to derive the BLEU.
% \xj{add the hyperparameter for bam here: learning rate, k, and say beta does not matter}
% \sz{working}

\begin{table}[h] %[8]{r}{0.4\textwidth} 
\vspace{-3.5mm}\centering
\caption{\small{Step time (sec) and number of parameters of variational attention [26] and BAM on NMT. }}\label{tb:cost}
\vspace{-1mm}
%\begin{sc}\vspace{-2mm}
\resizebox{0.32\columnwidth}{!}{\centering\begin{tabular}{ccc}\\\toprule  
& s/step & params  \\\midrule
% Soft attention & \bf{7.5e-2} & \bf{42M}\\
VA-Enum& 0.12& 64M\\
VA-Sample & 0.15& 64M\\
BAM-WC & \bf{0.10}& \bf{42M}\\
 \bottomrule
\end{tabular}}%\vspace{-6mm}
%\end{sc} 
\end{table}

% Model_soft_k5_6em4_beta1em6_acc_66.06_ppl_6.29_e15.pt
% |Param|: 1785.9287109375
% ELBO_q
% Loading valid dataset from data/iwslt/iwslt_125_test.valid.1.pt, number of examples: 6750
% Validation exp(elbo): 6.92636
% Validation perplexity: 6.92636
% Validation xent: 1.93534
% Validation kl: 0
% Validation accuracy: 64.5112
% N validation words: 152251

% BLEU = 33.81, 66.9/41.1/27.2/18.5 (BP=0.987, ratio=0.987, hyp_len=129401, ref_len=131141)

% \subsubsection{Ablation Study: The Impact of the Shape of Weibull Distribution}
% \xj{did you test this?}
% \sz{I have the final model saved but did not test as we discussed we would not need this anymore}
% \xj{I think we can still test this and check the performance. I still think this is important. }
% \label{sec:nmt_2}
% \begin{table}[htp!]\centering
% %\small
% \begin{sc}
% \caption{\small BLEU Under Different Shape K value.}
% \label{table:shapek}\resizebox{0.5\columnwidth}{!}{
% \begin{tabular}{@{}l|llll@{}}\toprule
% Shape (K) & 1 & 10 & 100 & 1000\\ \midrule
% BAM-WC  &  & & & \\ %(57812491)
% % Gaussian Dropout & 267K & 36.5M & 58M\\ %266610, 36541364
% % Concrete Dropout & 268K & 36.5M & 58M \\ %267794, ,57812553
% % Contextual & 311K & 36.6M& 61M\\%410098(10)/311426(32), 36622554 ,(61527401)
% \bottomrule
% \end{tabular}} %\vspace{-5mm}
% \end{sc}
% \end{table}

\subsection{Pretrained language models}\label{sec:app_plm}
\subsubsection{Model descriptions}
BERT \cite{devlin2018bert} is a state-of-the-art deep bidirectional transformer\cite{vaswani2017attention} model pretrained on large corpora to extract contextual word representations. ALBERT \cite{lan2019albert} improves upon BERT in terms of latency efficiency and performance by using (a) factorized embedding parameterization, (b) cross-layer parameter sharing, and (c) a sentence-order prediction (SOP) loss. Our experiment is done on the ALBERT-base model, which includes $12$ attention layers, each of hidden dimension $768$. The embedding dimension for factorized embedding is $128$. While BERT-base involves $108M$ parameters, ALBERT-base only has $12M$ parameters.
%  ALBERT denotes the vocabulary embedding size as E, the number of encoder
% layers as L, and the hidden size as H \cite{lan2019albert}. There are three distinct contribution of ALBERT 

\subsubsection{Detailed experimental settings}
Our experiment includes both the General Language Understanding Evaluation (GLUE) and Stanford Question Answering (SQuAD) Datasets. We evaluate on $8$ tasks from GLUE including Corpus of Linguistic Acceptability (CoLA; \cite{warstadt2019neural}), Stanford Sentiment Treebank (SST; \cite{socher2013recursive}), Microsoft Research Paraphrase Corpus (MRPC; \cite{dolan2005automatically}), Semantic Textual Similarity Benchmark (STS;\cite{cer2017semeval}), Quora Question Pairs (QQP; \cite{iyer2017first}), Multi-Genre NLI (MNLI; \cite{williams2017broad}), Question NLI (QNLI; \cite{rajpurkar2016squad}), and Recognizing Textual Entailment (RTE; \cite{dagan2005pascal}). We evaluate on both SQuAD v1.1 and SQuAD v2.0.
Our code is built on \citet{wolf2019transformers}, which can be found at \url{https://github.com/huggingface/transformers}. We follow the training settings as in \citet{lan2019albert} and summarize them in Table~\ref{tab:albert_setting}. We also include the hyperparameter setting for BAM-WC. We note, as the model is already pretrained so we do not anneal KL term. We pick $\beta=1\mathrm{E}\minus2$ and $d_\text{dim}=5$ for all experiments, as we found the performance is not sensitive to them. We include the $k$ in Table~\ref{tab:albert_setting}.
% is an extractive question answering dataset built from Wikipedia. The dataset are composed of two parts the answers (answers are derived from context paragraphs) and the task (the task is to predict answer spans). 
% For SQuAD v1.1, it has 100,000 human-annotated question/answer pairs and for SQuAD v2.0, it additionally introduced 50,000 unanswerable questions. Both SQuAD v1.1 and SQuAD v2.0 are pretrained under ALBERT \cite{lan2019albert}.

\begin{table}[htp!]\vspace{-5mm}
\caption{Experiment setting for pretrained language model (LR: learning rate, BSZ: batch size, DR: dropout rate, TS: training steps, WS: warmping steps, MSL: maximum sentence length).}\vspace{-1mm}
\label{tab:albert_setting}
\begin{center}
\begin{small}
\begin{sc}
\resizebox{0.9\columnwidth}{!}{
\begin{tabular}{@{}ccccccccc@{}}\toprule
& \text { LR } & \text { BSZ } & \text { ALBERT DR } & \text { Classifier DR } & \text { TS } & \text { WS } & \text { MSL } & $k$ \\ \midrule
\text { CoLA } & 1.00 $\mathrm{E}$\minus05 & 16 & 0 & 0.1 & 5336 & 320 & 512 & 10\\
\text { STS } & 2.00 $\mathrm{E}$\minus05 & 16 & 0 & 0.1 & 3598 & 214 & 512 & 20\\
\text { SST\minus2 } & 1.00 $\mathrm{E}$\minus05 & 32 & 0 & 0.1 & 20935 & 1256 & 512 &1000\\
\text { MNLI } & 3.00 $\mathrm{E}$\minus05 & 128 & 0 & 0.1 & 10000 & 1000 & 512 & 5\\
\text { QNLI } & 1.00 $\mathrm{E}$\minus05 & 32 & 0 & 0.1 & 33112 & 1986 & 512 & 500\\
\text { QQP } & 5.00 $\mathrm{E}$\minus05 & 128 & 0.1 & 0.1 & 14000 & 1000 & 512& 1000\\
\text { RTE } & 3.00 $\mathrm{E}$\minus05 & 32 & 0.1 & 0.1 & 800 & 200 & 512& 1000 \\
\text { MRPC } & 2.00 $\mathrm{E}$\minus05 & 32 & 0 & 0.1 & 800 & 200 & 512& 100\\
% \text { WNLI } & 2.00 $\mathrm{E}$\minus05 & 16 & 0.1 & 0.1 & 2000 & 250 & 512 \\
\text { SQuAD v1.1 } & 5.00 $\mathrm{E}$\minus05 & 48 & 0 & 0.1 & 3649 & 365 & 384& 10 \\
\text { SQuAD } v 2.0 & 3.00 $\mathrm{E}$\minus05 & 48 & 0 & 0.1 & 8144 & 814 & 512& 2000\\
\bottomrule
\end{tabular}}
\end{sc}
\end{small}
\end{center}
\vspace{-3mm}
\end{table}

\end{document}